\documentclass[preprint,12pt]{elsarticle}

\usepackage{times}
\usepackage{xcolor}
\usepackage{soul}
\usepackage[utf8]{inputenc}
\usepackage[small]{caption}

\usepackage[linesnumbered,ruled]{algorithm2e}
\usepackage{graphicx,amssymb,mathrsfs,amsmath,pifont,amscd,latexsym,amsfonts,amsthm}
\usepackage{graphics}
\usepackage{epstopdf}
\usepackage{float}
\usepackage[scriptsize]{subfigure}
\usepackage{epsfig}
\usepackage{amsfonts}
\usepackage{setspace}
\usepackage{multirow}
\newcommand{\norm}[1]{\lVert#1\rVert}
\usepackage{algpseudocode}
\usepackage{rotating}
\DeclareMathOperator{\F}{F}
\DeclareMathOperator{\T}{T}
\DeclareMathOperator{\prox}{prox}
\DeclareMathOperator{\sign}{sign}
\DeclareMathOperator{\svd}{svd}
\DeclareMathOperator{\tr}{tr}
\newcommand{\maxIter}{\textit{maxIter}}

\makeatletter
\def\hlinew#1{%
  \noalign{\ifnum0=`}\fi\hrule \@height #1 \futurelet
   \reserved@a\@xhline}
\journal{ISPRS}

\begin{document}

\begin{frontmatter}

\title{Learnable Manifold Alignment (LeMA) : A Semi-supervised Cross-modality Learning Framework for Land Cover and Land Use Classification}

 \author[label1,label2]{Danfeng Hong}
 \author[label3]{Naoto Yokoya}
 \author[label1]{Nan Ge}
 \author[label4]{Jocelyn Chanussot}
 \author[label1,label2]{Xiao Xiang Zhu}

 \address[label1]{Remote Sensing Technology Institute (IMF), German Aerospace Center (DLR), Wessling, Germany}
 \address[label2]{Signal Processing in Earth Observation (SiPEO), Technical University of Munich (TUM), Munich, Germany}
 \address[label3]{Geoinformatics Unit, RIKEN Center for Advanced Intelligence Project (AIP), RIKEN, Tokyo, Japan}
 \address[label4]{Univ. Grenoble Alpes, CNRS, Grenoble INP, GIPSA-lab, Grenoble, France}
\begin{abstract}
In this paper, we aim at tackling a general but interesting cross-modality feature learning question in remote sensing community --- \emph{can a limited amount of highly-discrimin-ative (e.g., hyperspectral) training data improve the performance of a classification task using a large amount of poorly-discriminative (e.g., multispectral) data?} Traditional semi-supervised manifold alignment methods do not perform sufficiently well for such problems, since the hyperspectral data is very expensive to be largely collected in a trade-off between time and efficiency, compared to the multispectral data. To this end, we propose a novel semi-supervised cross-modality learning framework, called learnable manifold alignment (LeMA). LeMA learns a joint graph structure directly from the data instead of using a given fixed graph defined by a Gaussian kernel function. With the learned graph, we can further capture the data distribution by graph-based label propagation, which enables finding a more accurate decision boundary. Additionally, an optimization strategy based on the alternating direction method of multipliers (ADMM) is designed to solve the proposed model. Extensive experiments on two hyperspectral-multispectral datasets demonstrate the superiority and effectiveness of the proposed method in comparison with several state-of-the-art methods.
\end{abstract}

\begin{keyword}


Cross-modality \sep graph learning \sep hyperspectral \sep manifold alignment \sep multispectral \sep remote sensing \sep semi-supervised learning.
\end{keyword}

\end{frontmatter}
\graphicspath{{Figure/}}

\section{Introduction}
Multispectral (MS) imagery has been receiving an increasing interest in the urban area (e.g. a large-scale land-cover mapping~\cite{huang2014multi}~\cite{hong2016k}, building localization~\cite{kang2018building}), agriculture~\cite{yang2013Agriculture}, and mineral products~\cite{van2014potential}, as operational optical broadband (multispectral) satellites (e.g. Sentinel-2 and Landsat-8 ~\cite{Naoto}) enable the multispectral imagery openly available on a global scale. In general, a reliable classifier needs to be trained on a large amount of labeled, discriminative, and high-quality samples. Unfortunately, labeling data, in particular large-scale data, is very gruelling and time-consuming. A natural alternative way to this issue is to consider tons of unlabeled data, yielding a semi-supervised learning. On the other hand, MS data fails to spectrally discriminate similar classes due to its broad spectral bandwidth. A simple way is to improve the data quality by fusing high-discriminative hyperspectral (HS) data~\cite{Naoto}. Although such data is expensive to collect, we may be able to expect a small amount of such data available. The aforementioned two points motivate us to raise a question related to transfer learning and cross-modality learning: \textit{Can a limited amount of HS training data partially overlapping MS data improve the performance of a classification task using a large coverage of MS testing data?}

Over the past decades, land-cover and land-use classification tasks of optical remote sensing imagery has received increasing attention in the unsupervised~\cite{hong2017DR}~\cite{li2014column} ~\cite{tarabalka2009spectral}, supervised~\cite{zhang2012combining}~\cite{hong2018joint}, and semi-supervised ways~\cite{xia2014semi}~\cite{Tuia}. To our best knowledge, the classifying ability in unsupervised learning (or dimensionality reduction) still remains limited, due to missing label information. By fully considering the variability of intra-class and inter-class from labels, supervised learning is able to perform the classification task better. In reality, a limited number of labeled samples usually hinders the trained classier towards a high classification performance, further leading to a possible failure in some challenging classification or transferring tasks owing to the lack of generalization and representability. Alternatively, semi-supervised learning draws into plenty of unlabeled data in learning process. This is capable of better capturing the distribution of different categories in order to find an accurate decision boundary.

On the other hand, considerable work related to transfer learning (TL) or domain adaptation (DA) has been successfully developed and applied in the remote sensing community~\cite{bruzzone2010domain,banerjee2015novel,matasci2015semisupervised,tuia2016domain,samat2016geodesic,samat2017supervised}. According to the different transferred objects, the TL or DA approaches can be roughly categorized into three groups, including parameter adaptation, instance-based transfer, and feature-based alignment or representation.

The seminal work dealing with parameter adaptation was presented in \cite{khosla2012undoing} and \cite{woodcock2001monitoring}, aiming at transferring an existing classifier (or parameters) trained or learned from the source domain to the target domain. Differently, the instance-based transferring technique transfers the knowledge by reweighting \cite{jiang2007instance} or resampling \cite{sugiyama2008direct} the samples of the source domain to those of the target domain. A similar idea based on active learning \cite{samat2016jointly} has also been proposed to address this issue, by selecting the most informative samples in the target domain to replace with those samples of the source domain that do not match the data distribution of the target domain \cite{persello2012active}.

For the final group of feature-based alignment or representation, manifold alignment (MA) is one of the most popular semi-supervised learning framework ~\cite{Wang:book} that facilitates transfer learning. MA has been successfully applied to various tasks in remote sensing community, e.g. classification \cite{tuia2016multi}, data visualization ~\cite{liao2016manifold}, multi-modality data analysis~\cite{Tuia}, etc.  The key idea of MA can be generalized as learning a common (or shared) subspace where different data can be aligned to learn a joint feature representation. Generally, existing MA methods can be approximately categorized into unsupervised, supervised, and semi-supervised approaches. The unsupervised approach usually fails to align multimodal data sufficiently well, as their corresponding low-dimensional embeddings may be quite diverse~\cite{Wang:AAAI2009}. In the supervised case, only aligning the limited number of training samples to learn a common subspace leads to weak transferability. While preserving a joint manifold structure created by both labeled and unlabeled data, semi-supervised alignment allows different data sources to be better transformed into the common subspace ~\cite{Wang:IJCAI2011}.

Although the joint manifold structure used in conventional semi-supervised MA approaches can relate features or instances, poor connections between the common subspace and label information still hinder the low-dimensional feature representation from being more discriminative. More importantly, in most graph-based semi-supervised learning algorithms (e.g. graph-based label propagation (GLP)~\cite{zhu2003semi}, semi-supervised manifold alignment (S-SMA~\cite{Tuia})~\cite{Wang:IJCAI2011}), the topology of unlabeled samples is merely given by a fixed Gaussian kernel function, which is computed in the original space rather than in the common space. This makes it difficult to adaptively transfer unlabeled samples into the learned common subspace, particularly when applied to multimodal data due to different numbers of dimensions. To address these issues, we propose a learnable manifold alignment (LeMA) by a data-driven graph learning directly from a common subspace so as to make the multimodal data comparable as well as improve the explainability of the learned common subspace, which further results in a better transferability. More specifically, our contributions can be summarized as follows:
\begin{itemize}
\item We propose a novel semi-supervised cross-modality learning framework called learnable manifold alignment (LeMA) for a large-scale land-cover classification task. One spectrally-poor MS and one spectrally rich HS data are considered as two different modalities and applied for this task, where the spatial extent of the former is a true superset of that of the latter.
\item Unlike jointly feature learning in which the model is both trained and tested from completed HS-MS correspondences, LeMA learns an aligned feature subspace from the labeled HS-MS correspondences and partially unlabeled MS data, and allows to identify out-of-samples using either MS data or HS data; Such the learnt subspace is a good fit for our case of cross-modality learning \footnote{In contrast to multi-modal learning (bi-modality for example), cross-modal learning trains on single modality and tests on bi-modality, or \emph{vice versa }(train on bi-modality and test on single modality).}.
\item Instead of directly computing graph structure with a Gaussian kernel function, a data-driven graph learning method is exploited behind LeMA in order to strengthen the abilities of transferring and generalization;
\item An optimization framework based on the alternating direction method of multipliers (ADMM) is designed to fast and effectively solve the proposed model.
\end{itemize}

The remainder of this paper is organized as follows. Section \uppercase\expandafter{\romannumeral2} elaborates on our motivation and proposes the methodology for the LeMA and the corresponding optimization algorithm. In Section \uppercase\expandafter{\romannumeral3}, we present the experimental results on two HS-MS datasets over the areas of the University of Houston and Chikusei, respectively,  and meanwhile discuss the qualitative and quantitative analysis. Section \uppercase\expandafter{\romannumeral4} concludes with a summary.
\begin{figure}[!t]
\centering
\includegraphics[width=1\textwidth]{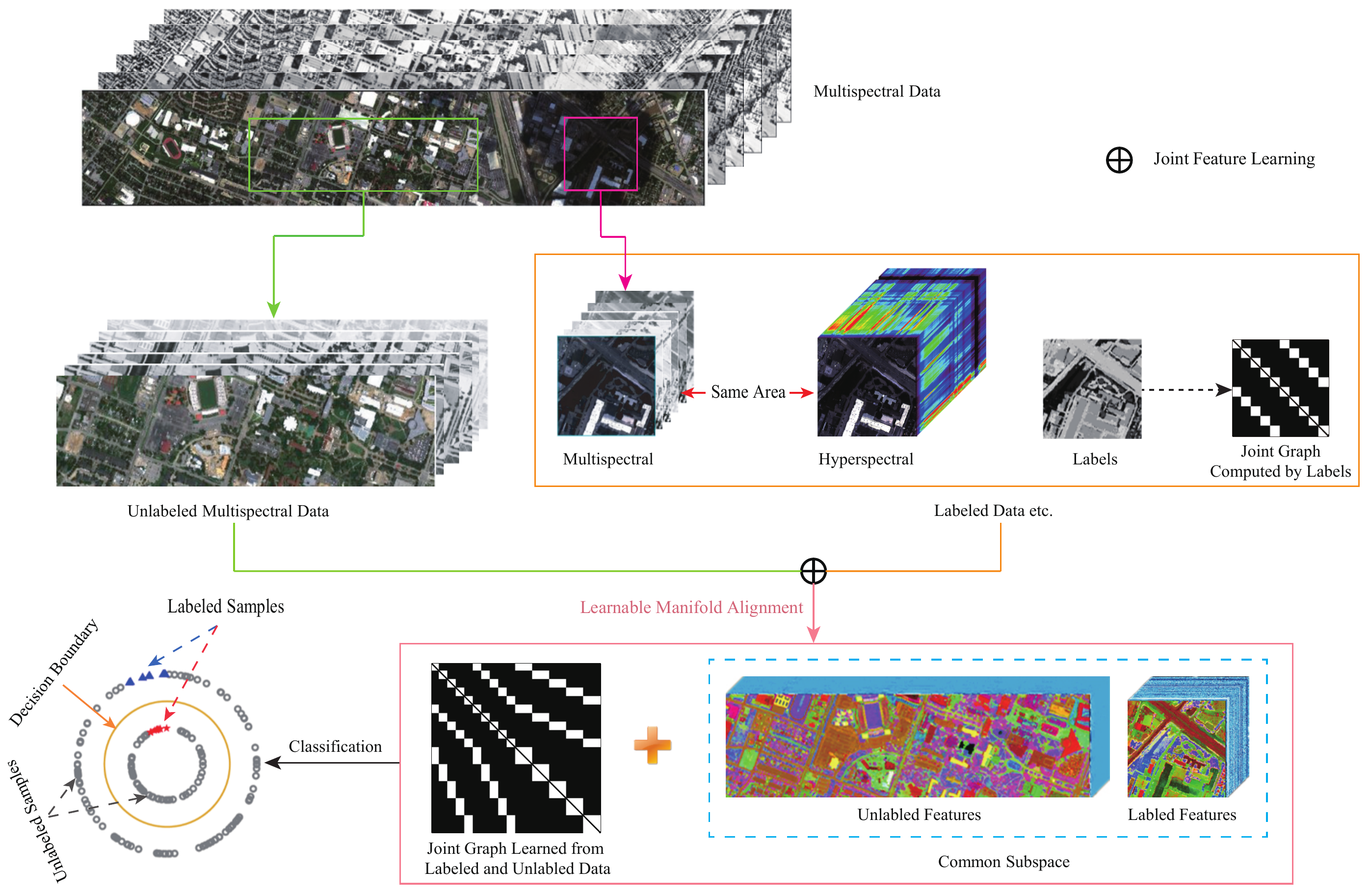}
\caption{An illustration of the proposed LeMA method.}
\label{fig:Workflow}
\end{figure}
\section{Learnable Manifold Alignment (LeMA)}
In this section, a cross-modality learning problem is firstly casted and the motivation is stated in the following. Accordingly, we formulate the methodology of our proposed and then elucidate an ADMM-based optimization algorithm to solve it.
\subsection{Problem Statement and Motivation}
For many high-level data analysis tasks in remote sensing community, such as land-cover classification, data collection plays an important role, since information-rich training samples enable us to easily find an optimal decision boundary.

There is, however, a typical bottleneck in collecting a large amount of labeled and discriminative data. Despite the MS data available at a global scale from the satellites of Sentinel-2 and Landsat-8, the identification and discrimination of materials are unattainable at an accuracy level by MS data, resulting from its poorly spectral information. On the contrary, HS data is characterized by rich spectral information, but only can be acquired in very small areas, due to the limitations of imaging sensors. This issue naturally guides us to jointly utilize the HS and MS bi-modal data, specifically leading to the following interesting and challenging question \emph{can a limited number of HS training data contribute to the classification task of a large-scale MS data?}

A feasible solution to the issue can be unfolded to two parts: 1) \emph{cross-modality learning}: learning a common subspace where the features are expected to absorb the different properties from the HS-MS modalities and meanwhile the HS and MS data can be transferred each other; 2) \emph{semi-supervised learning}: Embedding massive unlabeled MS samples which are relatively in large quantities and easy to be collected, so as to learn a more discriminative feature representation. Fig.~\ref{fig:Workflow} illustrates the workflow of LeMA.
\begin{figure}[!t]
\centering
\includegraphics[width=0.5\textwidth]{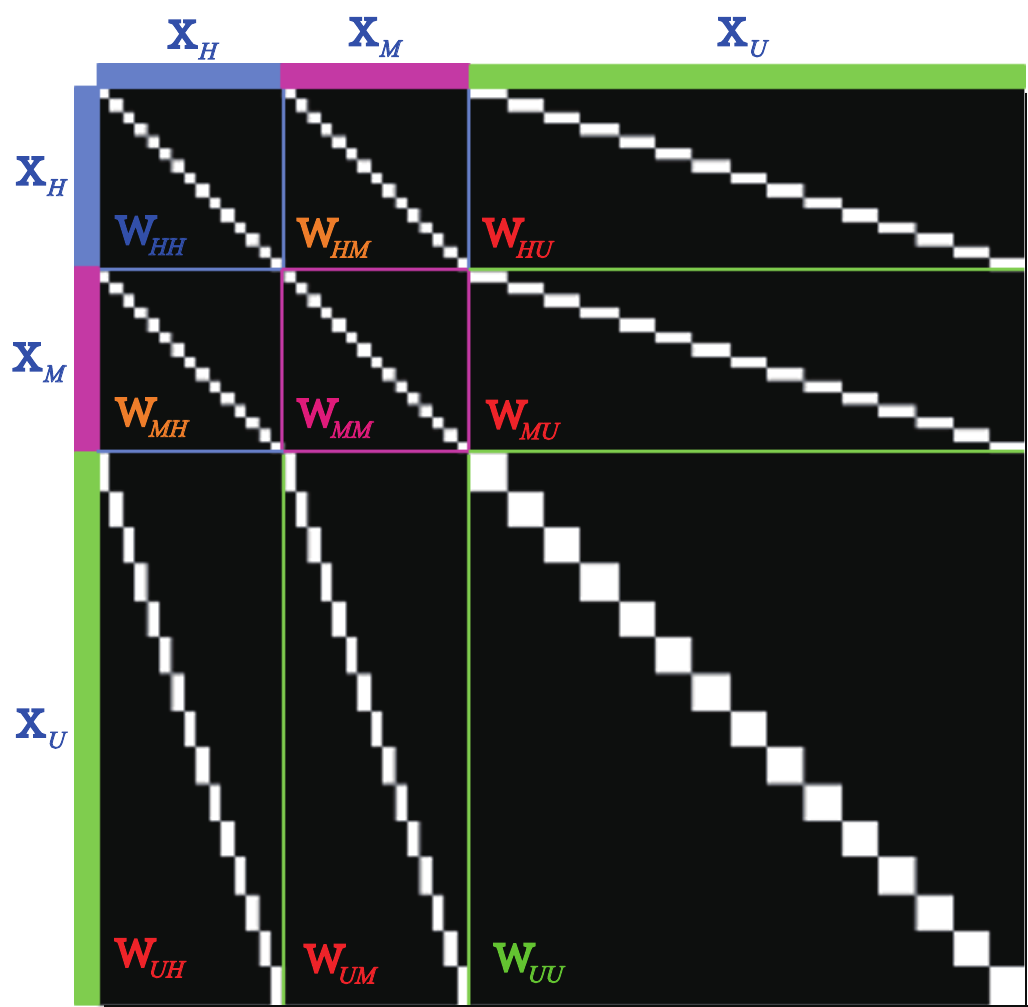}
\caption{An example for the joint adjacency matrix $\widetilde{\mathbf{W}}$.}
\label{Fig2}
\end{figure}
\subsection{Problem Formulation}
To effectively model the aforementioned issue, we intend to develop a joint learning framework which better learns a discriminative common subspace from high-quality HS data and low-quality MS data. Intuitively, such a common subspace can be shaped by selectively absorbing the benefits of both high-quality data with more details and low-quality data with more structural information. Therefore, following a popular joint learning framework~\cite{Ji:IJCAI2009}, we formulate the common subspace learning problem as
\begin{equation}
\label{eq1}
\begin{aligned}
       \mathop{\min}_{\mathbf{P},\mathbf{\Theta}}\; \frac{1}{2}\norm{\widetilde{\mathbf{Y}}-\mathbf{P}\mathbf{\Theta}\widetilde{\mathbf{X}}}_{\F}^{2}+\frac{\alpha}{2}\norm{\mathbf{P}}_{\F}^{2}+\frac{\beta}{2}\tr(\mathbf{E}\mathbf{L}\mathbf{E}^{\T})\;
       \mathrm{s.t.} \; \mathbf{E}=\mathbf{\Theta}\widetilde{\mathbf{X}}, \; \mathbf{\Theta}\mathbf{\Theta}^{\T}=\mathbf{I},
\end{aligned}
\end{equation}
where $\widetilde{\mathbf{Y}}=\left[\mathbf{Y},\mathbf{Y}\right]\in\mathbb{R}^{d\times 2N}$ and $\mathbf{Y}\in\mathbb{R}^{d\times N}$ is the label matrix represented by one-hot encoding,
$\widetilde{\mathbf{X}}=
\begin{bmatrix}
     \mathbf{X}_{H} & \mathbf{0} \\
     \mathbf{0} & \mathbf{X}_{M}
\end{bmatrix}\in\mathbb{R}^{(d_{H}+d_{M})\times 2N}$ and $\mathbf{X}_{H}$ and $\mathbf{X}_{M}$ stand respectively for the data from hyperspectral and multispectral domains, $\mathbf{\Theta}=\left[\mathbf{\Theta}_{H},\mathbf{\Theta}_{M}\right]$ and $\mathbf{P}$ are respectively the common subspace projection and the linear projection to bridge the common subspace and label information. $\mathbf{L}=\mathbf{D}-\mathbf{W}\in\mathbb{R}^{2N\times 2N}$ stands for a joint Laplacian matrix, $\mathbf{W}$ is an adjacency matrix and $\mathbf{D}_{ii}=\sum_{i\neq j}\mathbf{W}_{i,j}$. $\mathbf{W}$ is generally used to measure the similarity between samples. With the orthogonal constraint ($\mathbf{\Theta}\mathbf{\Theta}^{\T}=\mathbf{I}$), the global optimal solutions with respect to the variables $\mathbf{\Theta}$ and $\mathbf{P}$ can be theoretically guaranteed~\cite{Ji:IJCAI2009}.
\begin{algorithm}[!t]
\footnotesize
\label{alg1}
\caption{Learnable Manifold Alignment (LeMA)}
\KwIn{$\widetilde{\mathbf{Y}}$, $\widetilde{\mathbf{X}}$, $\widetilde{\mathbf{X}}'$, $\widetilde{\mathbf{L}}$, $\alpha$, $\beta$, $\maxIter$.}
\KwOut{$\mathbf{P},\mathbf{\Theta},\widetilde{\mathbf{L}}$}
 $t=1$, $\zeta=1e-4$;\\
{\textbf{Initializating} $\mathbf{P}$ and $\mathbf{\Theta}$}\\

  \While{not converged \rm{or} $t>\maxIter$}
 {
   Fix other variables to update $\mathbf{P}$ by Eq.~(\ref{eq6})

   Fix other variables to update $\mathbf{\Theta}$ by \textbf{Algorithm} $\mathbf{2}$

   Fix other variables to update $\widetilde{\mathbf{L}}$ by equivalently optimizing $\widetilde{\mathbf{W}}$ in a distributed fashion:

   \quad 1. update $\widetilde{\mathbf{W}}_{HU}$ by \textbf{Algorithm} $\mathbf{3}$;

   \quad 2. update $\widetilde{\mathbf{W}}_{MU}$ by \textbf{Algorithm} $\mathbf{3}$;

   \quad 3. align $\widetilde{\mathbf{W}}_{HU}$ and $\widetilde{\mathbf{W}}_{MU}$ by $\max(\widetilde{\mathbf{W}}_{HU},\widetilde{\mathbf{W}}_{MU})$;

   \quad 4. update $\widetilde{\mathbf{W}}_{UU}$ by \textbf{Algorithm} $\mathbf{4}$

   \quad 5. compute $\widetilde{\mathbf{L}}=\widetilde{\mathbf{D}}-\widetilde{\mathbf{W}}$, $\widetilde{\mathbf{D}}_{ii}=\sum_{i\neq j}\widetilde{\mathbf{W}}_{ij}$

   Compute the objective function value $E^{t+1}$ and check the convergence condition:
   \eIf{$|\frac{E^{t+1}-E^{t}}{E^{t}}|<\zeta$}
   {
     Stop iteration;
   }
   {
     $t\leftarrow t+1$;
   }
 }
\end{algorithm}

The first term of Eq.~(\ref{eq1}) is a fidelity term, and the regularization term $\frac{\alpha}{2}\norm{\mathbf{P}}_{\F}^{2}$ parameterized by $\alpha$ aims to achieve a reliable generalization of the proposed model. The third term acts as supervised manifold alignment (SMA) ~\cite{Wang:book}. We refer to the proposed framework for joint common subspace learning as CoSpace.

To further exploit the information of unlabeled samples, we extend the CoSpace in Eq.~(\ref{eq1}) to LeMA by learning a joint Laplacian matrix, which can be formulated as follows with extra constraints related to necessary conditions of $\widetilde{\mathbf{L}}$:
\begin{equation}
\label{eq2}
\begin{aligned}
       &\mathop{\min}_{\mathbf{P},\mathbf{\Theta},\widetilde{\mathbf{L}}}\; \frac{1}{2}\norm{\widetilde{\mathbf{Y}}-\mathbf{P}\mathbf{\Theta}\widetilde{\mathbf{X}}}_{\F}^{2}+\frac{\alpha}{2}\norm{\mathbf{P}}_{\F}^{2}+\frac{\beta}{2}\tr(\mathbf{H}\widetilde{\mathbf{L}}\mathbf{H}^{\T})\\
       &\mathrm{s.t.} \; \mathbf{H}=\mathbf{\Theta}\widetilde{\mathbf{X}}', \; \mathbf{\Theta}\mathbf{\Theta}^{\T}=\mathbf{I},\; \widetilde{\mathbf{L}}=\widetilde{\mathbf{L}}^{\T},\; \widetilde{\mathbf{L}}_{i,j,i\neq j}\preceq 0, \; \widetilde{\mathbf{L}}_{i,j,i=j}\succeq 0, \; \tr(\widetilde{\mathbf{L}})=s,
\end{aligned}
\end{equation}
where $\widetilde{\mathbf{X}}'=
\begin{bmatrix}
     \mathbf{X}_{H} & \mathbf{0} & \mathbf{0}\\
     \mathbf{0} & \mathbf{X}_{M} & \mathbf{X}_{U}
\end{bmatrix}\in\mathbb{R}^{(d_{H}+d_{M})\times (2N+N_{U})}$, $\widetilde{\mathbf{L}}\in\mathbb{R}^{(2N+N_{U})\times(2N+N_{U})}$, and $\mathbf{X}_{U}\in\mathbb{R}^{d_{M}\times N_{U}}$ represents the unlabeled MS samples and $s> 0$ controls the scale. Note that a feasible and effective approach to choose the unlabeled data with respect to the variable $\widetilde{\mathbf{X}}'$ is to group total samples besides the training samples into some landmarks (cluster centers). These landmarks are used as the unlabeled data, which can fully take into account the available information and meanwhile effectively reduce the computational cost. Due to the use of clustering technique in unlabeled data, we experimentally and empirically set the ratio of labeled and unlabeled data to approximately be 1:1.

The model in Eq. (\ref{eq2}) can be simplified by optimizing the adjacency matrix ($\widetilde{\mathbf{W}}$) instead of directly solving a hard optimization problem of $\widetilde{\mathbf{L}}$, then we have
\begin{equation}
\label{eq3}
\begin{aligned}
     \tr(\mathbf{H}\widetilde{\mathbf{L}}\mathbf{H}^{\T})=\frac{1}{2}\tr(\widetilde{\mathbf{W}}\mathbf{Z})=\frac{1}{2}\norm{\widetilde{\mathbf{W}}\odot\mathbf{Z}}_{1,1},
\end{aligned}
\end{equation}
where $\widetilde{\mathbf{W}}\in\mathbb{R}^{(2N+N_{U})\times (2N+N_{U})}$, $\mathbf{Z}\in\mathbb{R}^{(2N+N_{U})\times (2N+N_{U})}$ is defined as {\it a pairwise Euclidean distance matrix} : $\mathbf{Z}_{i,j}=\norm{\mathbf{H}_{i}-\mathbf{H}_{j}}^{2}$. $\odot$ denotes the Schur-Hadamard (termwise) product.
\begin{algorithm}[!t]
\footnotesize
\caption{Solving the subproblem for $\mathbf{\Theta}$}
\KwIn{$\widetilde{\mathbf{Y}}$, $\mathbf{P}$, $\mathbf{J}$, $\widetilde{\mathbf{X}}$, $\widetilde{\mathbf{X}}'$, $\widetilde{\mathbf{L}}$, $\beta$, $\maxIter$.}
\KwOut{$\mathbf{\Theta}.$}
\textbf{Initialization}:
$\mathbf{\Theta}=\mathbf{0}$, $\mathbf{G}=\mathbf{0}$, $\mathbf{\Lambda}_{1}=\mathbf{0}$,
$\mathbf{\Lambda}_{2}=\mathbf{0}$, $\mu=10^{-3}$, $\mu_{\max}=10^{6}$, $\rho=1.5$, $\varepsilon=10^{-6}$, $t=1$.\\
\While{not converged \rm{or} $t>\maxIter$}
 {
         Fix other variables to update $\mathbf{J}$ by $\mathbf{J}=(\mathbf{P}^{\T}\mathbf{P}+\mu\mathbf{I})^{-1}(\mathbf{P}^{\T}\widetilde{\mathbf{Y}}+\mu\mathbf{\Theta}\widetilde{\mathbf{X}}-\mathbf{\Lambda}_{1}).$\\
         Fix other variables to update $\mathbf{\Theta}$ by
         {\setlength\abovedisplayskip{1pt}
         \setlength\belowdisplayskip{1pt}
         \begin{equation*}
         \begin{aligned}
            \mathbf{\Theta}=(\mu\mathbf{J}\widetilde{\mathbf{X}}^{\T}+\mathbf{\Lambda}_{1}\widetilde{\mathbf{X}}^{\T}+\mu\mathbf{G}+\mathbf{\Lambda}_{2})
             \times(\mu\widetilde{\mathbf{X}}\widetilde{\mathbf{X}}^{\T}+\mu\mathbf{I}+\beta\widetilde{\mathbf{X}}'\widetilde{\mathbf{L}}\widetilde{\mathbf{X}}'^{\T})^{-1}.
         \end{aligned}
         \end{equation*}}\\
         Fix other variables to update $\mathbf{G}$ by
         {\setlength\abovedisplayskip{1pt}
         \setlength\belowdisplayskip{1pt}
         \begin{equation*}
         \begin{aligned}
                \left[\mathbf{U},\mathbf{S},\mathbf{V}\right]=\svd(\mathbf{\mathbf{\Theta}-\mathbf{\Lambda}_{2}/\mu}), \quad \mathbf{G}=\mathbf{U}\mathbf{I}_{n \times m}\mathbf{V}.
         \end{aligned}
         \end{equation*}}\\
         Update Lagrange multipliers by
         {\setlength\abovedisplayskip{1pt}
         \setlength\belowdisplayskip{1pt}
         \begin{equation*}
         \begin{aligned}
                \mathbf{\Lambda}_{1} \leftarrow \mathbf{\Lambda}_{1}+\mu(\mathbf{J}-\mathbf{\Theta}\widetilde{\mathbf{X}}), \quad \mathbf{\Lambda}_{2} \leftarrow \mathbf{\Lambda}_{2}+\mu(\mathbf{G}-\mathbf{\Theta}).
         \end{aligned}
         \end{equation*}}\\
         Update penalty parameter by $\mu=\min (\rho\mu, \mu_{\max}).$\\
         Check the convergence conditions:
         \eIf{$\norm {\mathbf{J}-\mathbf{\Theta}\widetilde{\mathbf{X}}}_{\F}<\varepsilon$ and $\norm {\mathbf{G}-\mathbf{\Theta}}_{\F}<\varepsilon$}
         {
           Stop iteration;
         }
         {
         $t\leftarrow t+1$;
         }
 }
\end{algorithm}

Using Eq.~(\ref{eq3}), we can equivalently convert the optimization problem of smooth manifold in (\ref{eq2}) to that of graph sparsity
\begin{equation}
\label{eq4}
\begin{aligned}
       &\mathop{\min}_{\mathbf{P},\mathbf{\Theta},\widetilde{\mathbf{W}}} \; \frac{1}{2}\norm{\widetilde{\mathbf{Y}}-\mathbf{P}\mathbf{\Theta}\widetilde{\mathbf{X}}}_{\F}^{2}+\frac{\alpha}{2}\norm{\mathbf{P}}_{\F}^{2}+\frac{\beta}{4}\norm{\widetilde{\mathbf{W}}\odot\mathbf{Z}}_{1,1}\\
       &\mathrm{s.t.} \; \mathbf{H}=\mathbf{\Theta}\widetilde{\mathbf{X}}', \; \mathbf{\Theta}\mathbf{\Theta}^{\T}=\mathbf{I},\; \widetilde{\mathbf{W}}=\widetilde{\mathbf{W}}^{\T}, \; \widetilde{\mathbf{W}}_{i,j}\succeq 0, \; \norm{\widetilde{\mathbf{W}}}_{1,1}=s,
\end{aligned}
\end{equation}
where $\norm{\widetilde{\mathbf{W}}\odot\mathbf{Z}}_{1,1}$ can be interpreted as a \emph{weighted $\ell_{1}$-norm of $\widetilde{\mathbf{W}}$ which enforces weighted sparsity}.

We further elaborate the relationship between the proposed LeMA model and our motivation in an easy-understanding way. In general, we aim at finding a common subspace by learning a pair of projections ($\mathbf{\Theta}_{M}$ and $\mathbf{\Theta}_{H}$) corresponding to two kinds of different modalities (e.g., MS and HS), respectively. In order to effectively improve the discriminative ability of the learned subspace, we make a connection between the subspace and label information by jointly estimating the regression coefficient $\mathbf{P}$ and common projections $\mathbf{\Theta}$, as formulated in Eq. (\ref{eq1}). What's more, the alignment behavior of different modalities can be represented by $\mathbf{W}$'s connectivity, that is, if the $i^{th}$ sample $\mathbf{X}_{i}$ and the $j^{th}$ sample $\mathbf{X}_{j}$ are connected ($\mathbf{W}_{i,j}=1$), and then the two samples belong to the same class; \textit{vice versa}. Besides, we construct an extra adjacency matrix based on those unlabeled samples in order to globally capture the data distribution. The matrix is usually obtained by a Gaussian kernel function (semi-supervised CoSpace) and also can be learned from the data (LeMA as formulated in Eq. (\ref{eq2})).
\begin{algorithm}[!t]
\footnotesize
\caption{Solving the subproblem for $\widetilde{\mathbf{W}}_{HU(MU)}$}
\KwIn{$\mathbf{Z}_{H(M)}$, $\mathbf{Z}_{U}$, $\widetilde{\mathbf{W}}$, $\beta$, $\maxIter$.}
\KwOut{$\widetilde{\mathbf{W}}.$}
\textbf{Initialization}:
$\mathbf{M}=\widetilde{\mathbf{W}}$, $\mathbf{S}=\mathbf{U}=\mathbf{K}=\mathbf{0}$, $\mathbf{\Lambda}_{1}=\mathbf{\Lambda}_{2}=\mathbf{\Lambda}_{3}=\mathbf{\Lambda}_{4}=\mathbf{0}$, $\mu=10^{-2}$, $\mu_{\max}=10^6$, $\rho=2$, $\varepsilon=10^{-6}$, $t=1$.\\
\textbf{Compute $\mathbf{Z}$}: $\mathbf{Z}_{i,j}=\norm{\mathbf{Z}_{H(M)}^{i}-\mathbf{Z}_{U}^{j}}_{\F}^{2}$.\\
\While{not converged \rm{or} $t>\maxIter$}
 {
         Fix other variables to update $\widetilde{\mathbf{W}}$ by
         {\setlength\abovedisplayskip{1pt}
         \setlength\belowdisplayskip{1pt}
         \begin{equation*}
         \begin{aligned}
              \widetilde{\mathbf{W}}=(\mathbf{M}+\mathbf{S}+\mathbf{U}+\mathbf{K}+\mathbf{\Lambda}_{1}+\mathbf{\Lambda}_{2}++\mathbf{\Lambda}_{3}+\mathbf{\Lambda}_{4})/(4\mu).
         \end{aligned}
         \end{equation*}}\\
         Fix other variables to update $\mathbf{U}$ by $ \mathbf{U}=\max(\widetilde{\mathbf{W}}-\mathbf{\Lambda}_{1}/\mu,0).$\\
         Fix other variables to update $\mathbf{M}$ by
         {\setlength\abovedisplayskip{1pt}
         \setlength\belowdisplayskip{1pt}
         \begin{equation*}
         \begin{aligned}
                \mathbf{M}=\max(\norm{\widetilde{\mathbf{W}}-\mathbf{\Lambda}_{2}/\mu}_{1,1}-(\beta\mathbf{Z}/4\mu),0)\odot \sign(\widetilde{\mathbf{W}}-\mathbf{\Lambda}_{2}/\mu).
         \end{aligned}
         \end{equation*}}\\
         Fix other variables to update $\mathbf{S}$ by $\mathbf{S}=\prox(\widetilde{\mathbf{W}}-\mathbf{\Lambda}_{3}/\mu).$\\
         Fix other variables to update $\mathbf{K}$ by $\mathbf{K}=\min(\widetilde{\mathbf{W}}-\mathbf{\Lambda}_{4}/\mu,1/N_{k}).$\\
         Update Lagrange multipliers by
         {\setlength\abovedisplayskip{1pt}
         \setlength\belowdisplayskip{1pt}
         \begin{equation*}
         \begin{aligned}
                \mathbf{\Lambda}_{1}&=\mathbf{\Lambda}_{1}+\mu(\mathbf{U}-\widetilde{\mathbf{W}}), \quad \mathbf{\Lambda}_{2}=\mathbf{\Lambda}_{2}+\mu(\mathbf{M}-\widetilde{\mathbf{W}}),\\
                \mathbf{\Lambda}_{3}&=\mathbf{\Lambda}_{3}+\mu(\mathbf{S}-\widetilde{\mathbf{W}}), \quad \mathbf{\Lambda}_{4}=\mathbf{\Lambda}_{4}+\mu(\mathbf{K}-\widetilde{\mathbf{W}}).
         \end{aligned}
         \end{equation*}}\\
         Update penalty parameter by $\mu=\min (\rho\mu,\mu_{\max}).$
         Check the convergence conditions:
         \eIf{$\norm {\mathbf{U}-\widetilde{\mathbf{W}}}_{\F}<\varepsilon$ and $\norm {\mathbf{M}-\widetilde{\mathbf{W}}}_{\F}<\varepsilon$ and $\norm {\mathbf{S}-\widetilde{\mathbf{W}}}_{\F}<\varepsilon$ and $\norm {\mathbf{K}-\widetilde{\mathbf{W}}}_{\F}<\varepsilon$ and $\norm {\widetilde{\mathbf{W}}^{t+1}-\widetilde{\mathbf{W}}^{t}}_{F}<\varepsilon$}
         {
           Stop iteration;
         }
         {
         $t\leftarrow t+1$;
         }
 }
 \end{algorithm}
\subsection{Model Optimization}
Considering the complexity of the non-convex problem (\ref{eq4}), an iterative alternating optimization strategy is adopted to solve the convex subproblems of each variable $\mathbf{P}$, $\mathbf{\Theta}$, and $\mathbf{W}$. An implementation of LeMA is given in \textbf{Algorithm~1}.

\emph{Optimization with respect to $\mathbf{P}$}: This is a typical least-squares problem with Tikhonov regularization, which can be formulated as
\begin{equation}
\label{eq5}
\begin{aligned}
       \mathop{\min}_{\mathbf{P}} \; \frac{1}{2}\norm{\widetilde{\mathbf{Y}}-\mathbf{P}\mathbf{\Theta}\widetilde{\mathbf{X}}}_{\F}^{2}+\frac{\alpha}{2}\norm{\mathbf{P}}_{\F}^{2},
\end{aligned}
\end{equation}
which has a closed-form solution
\begin{equation}
\label{eq6}
\begin{aligned}
        \mathbf{P} = (\widetilde{\mathbf{Y}}\mathbf{E}^{\T})(\mathbf{E}\mathbf{E}^{\T}+\alpha\mathbf{I})^{-1},
\end{aligned}
\end{equation}
where $\mathbf{E}=\mathbf{\Theta}\widetilde{\mathbf{X}}$.

\emph{Optimization with respect to $\mathbf{\Theta}$}: the optimization problem for $\mathbf{\Theta}$ can be formulated as
\begin{equation}
\label{eq7}
\begin{aligned}
       \mathop{\min}_{\mathbf{\Theta}} \; \frac{1}{2}\norm{\widetilde{\mathbf{Y}}-\mathbf{P}\mathbf{\Theta}\widetilde{\mathbf{X}}}_{\F}^{2}+\frac{\beta}{2}\tr(\mathbf{H}\widetilde{\mathbf{L}}\mathbf{H}^{\T})\;
       \mathrm{s.t.} \; \mathbf{H}=\mathbf{\Theta}\widetilde{\mathbf{X}}', \; \mathbf{\Theta}\mathbf{\Theta}^{\T}=\mathbf{I}.
\end{aligned}
\end{equation}
In order to solve (\ref{eq7}) effectively with ADMM, we consider an equivalent form by introducing auxiliary variables $\mathbf{J}$ and $\mathbf{G}$ to replace $\mathbf{\Theta}\widetilde{\mathbf{X}}$ and $\mathbf{\Theta}$, respectively.
\begin{equation}
\label{eq8}
\begin{aligned}
       &\mathop{\min}_{\mathbf{\Theta},\mathbf{J},\mathbf{G}}\; \frac{1}{2}\norm{\widetilde{\mathbf{Y}}-\mathbf{P}\mathbf{J}}_{\F}^{2}+\frac{\beta}{2}\tr(\mathbf{\Theta}\widetilde{\mathbf{X}}'\widetilde{\mathbf{L}}(\mathbf{\Theta}\widetilde{\mathbf{X}}')^{\T})\\
       &\mathrm{s.t.} \; \mathbf{J}=\mathbf{\Theta}\widetilde{\mathbf{X}}, \; \mathbf{G}=\mathbf{\Theta}, \; \mathbf{G}\mathbf{G}^{\T}=\mathbf{I}.
\end{aligned}
\end{equation}
\textbf{Algorithm~2} lists the more detailed procedures for solving the problem (\ref{eq8}).
\begin{algorithm}[!t]
\footnotesize
\caption{Solving the subproblem for $\widetilde{\mathbf{W}}_{UU}$}
\KwIn{$\mathbf{Z}_{U}$, $\widetilde{\mathbf{W}}$, $\gamma$, $\maxIter$.}
\KwOut{$\widetilde{\mathbf{W}}.$}
\textbf{Initialization}:
$\mathbf{M}=\widetilde{\mathbf{W}}$, $\mathbf{U}=\mathbf{V}=\mathbf{S}=\mathbf{K}=\mathbf{T}=\mathbf{0}$, $\mathbf{\Lambda}_{1}=\mathbf{\Lambda}_{2}=\mathbf{\Lambda}_{3}=\mathbf{\Lambda}_{4}=\mathbf{\Lambda}_{5}=\mathbf{\Lambda}_{6}=\mathbf{\Lambda}_{7}=\mathbf{0}$, $\mu=10^{-2}$, $\mu_{\max}=10^6$, $\rho=2$, $\varepsilon=10^{-6}$, $t=1$.\\
\textbf{Compute $\mathbf{Z}$}: $\mathbf{Z}_{i,j}=\norm{\mathbf{Z}_{U}^{i}-\mathbf{Z}_{U}^{j}}_{\F}^{2}$.\\
\While{not converged \rm{or} $t>\maxIter$}
 {
         Fix other variables to update $\widetilde{\mathbf{W}}$ by
         {\setlength\abovedisplayskip{1pt}
         \setlength\belowdisplayskip{1pt}
         \begin{equation*}
         \begin{aligned}
               \widetilde{\mathbf{W}}=(\mathbf{V}+\mathbf{U}^{\T}+\mathbf{M}+\mathbf{S}+\mathbf{K}+\mathbf{T}+\mathbf{\Lambda}_{1}+\mathbf{\Lambda}_{2}^{\T} +\mathbf{\Lambda}_{3}+\mathbf{\Lambda}_{4}+\mathbf{\Lambda}_{5}+\mathbf{\Lambda}_{7})/(6\mu).
         \end{aligned}
         \end{equation*}}\\
         Fix other variables to update $\mathbf{U}$ by $
                \mathbf{U}=\big(\widetilde{\mathbf{W}}^{\T}+\mathbf{V}-(\mathbf{\Lambda}_{1}+\mathbf{\Lambda}_{6})\big)/(2\mu).$\\
         Fix other variables to update $\mathbf{V}$ by $
                \mathbf{V}=\big(\widetilde{\mathbf{W}}+\mathbf{U}-(\mathbf{\Lambda}_{2}+\mathbf{\Lambda}_{6})\big)/(2\mu).$\\
         Fix other variables to update $\mathbf{M}$ by
         {\setlength\abovedisplayskip{1pt}
         \setlength\belowdisplayskip{1pt}
         \begin{equation*}
         \begin{aligned}
                \mathbf{M}=\max(\norm{\widetilde{\mathbf{W}}-\mathbf{\Lambda}_{3}/\mu}_{1,1}-\gamma\mathbf{Z}/(4\mu),0)\odot \sign(\widetilde{\mathbf{W}}-\mathbf{\Lambda}_{3}/\mu).
         \end{aligned}
         \end{equation*}}\\
         Fix other variables to update $\mathbf{S}$ by $\mathbf{S}=\prox (\widetilde{\mathbf{W}}-\mathbf{\Lambda}_{4}/\mu).$\\
         Fix other variables to update $\mathbf{K}$ by $\mathbf{K}=\max(\widetilde{\mathbf{W}}-\mathbf{\Lambda}_{5}/\mu,0).$\\
         Fix other variables to update $\mathbf{T}$ by
         $\mathbf{T}=\min(\widetilde{\mathbf{W}}-\mathbf{\Lambda}_{7}/\mu,1/N_{k}).$\\
         Update Lagrange multipliers by
         {\setlength\abovedisplayskip{1pt}
         \setlength\belowdisplayskip{1pt}
         \begin{equation*}
         \begin{aligned}
                \mathbf{\Lambda}_{1}&=\mathbf{\Lambda}_{1}+\mu(\mathbf{U}-\widetilde{\mathbf{W}}^{\T}), & \mathbf{\Lambda}_{2} & =\mathbf{\Lambda}_{2}+\mu(\mathbf{V}-\widetilde{\mathbf{W}}),\\
                \mathbf{\Lambda}_{3}&=\mathbf{\Lambda}_{3}+\mu(\mathbf{M}-\widetilde{\mathbf{W}}), & \mathbf{\Lambda}_{4} & =\mathbf{\Lambda}_{4}+\mu(\mathbf{S}-\widetilde{\mathbf{W}}),\\
                \mathbf{\Lambda}_{5}&=\mathbf{\Lambda}_{5}+\mu(\mathbf{K}-\widetilde{\mathbf{W}}), & \mathbf{\Lambda}_{6} & =\mathbf{\Lambda}_{6}+\mu(\mathbf{U}-\mathbf{V}),\\
                \mathbf{\Lambda}_{7}&=\mathbf{\Lambda}_{7}+\mu(\mathbf{T}-\widetilde{\mathbf{W}}).
         \end{aligned}
         \end{equation*}}\\
         Update penalty parameter by $\mu=\min (\rho\mu,\mu_{\max}).$\\
         Check the convergence conditions:
         \eIf{$\norm {\mathbf{U}-\widetilde{\mathbf{W}}^{\T}}_{\F}<\varepsilon$ and $\norm {\mathbf{V}-\widetilde{\mathbf{W}}}_{\F}<\varepsilon$ and $\norm {\mathbf{M}-\widetilde{\mathbf{W}}}_{\F}<\varepsilon$ and $\norm {\mathbf{S}-\widetilde{\mathbf{W}}}_{\F}<\varepsilon$ and $\norm {\mathbf{K}-\widetilde{\mathbf{W}}}_{\F}<\varepsilon$ and $\norm {\mathbf{U}-\mathbf{V}}_{\F}<\varepsilon$ and $\norm {\mathbf{T}-\widetilde{\mathbf{W}}}_{\F}<\varepsilon$ and $\norm {\widetilde{\mathbf{W}}^{t+1}-\widetilde{\mathbf{W}}^{t}}_{\F}<\varepsilon$}
         {
           Stop iteration;
         }
         {
         $t\leftarrow t+1$;
         }
 }
\end{algorithm}

\emph{Optimization with respect to $\widetilde{\mathbf{W}}$}: $\widetilde{\mathbf{W}}$ is a joint adjacency matrix and consists mainly of nine parts as shown in Fig.~\ref{Fig2}. Among the nine parts,  $\widetilde{\mathbf{W}}_{HH}$, $\widetilde{\mathbf{W}}_{HM}$, $\widetilde{\mathbf{W}}_{MH}$ and $\widetilde{\mathbf{W}}_{MM}$ can be directly inferred from label information in the form of the LDA-like graph ~\cite{Gu:IJCAI2011}:
\begin{equation}
\label{eq9}
  \widetilde{\mathbf{W}}_{i,j}=
    \begin{cases}
      \begin{aligned}
      1/N_k, \quad & \text{if \(\mathbf{X}_{i}\) and \(\mathbf{X}_{j}\) belong to the \(k\)-th class;}\\
      0, \quad & \text{otherwise.}
      \end{aligned}
    \end{cases}
\end{equation}
Given the symmetry of $\widetilde{\mathbf{W}}$, (i.e., $\widetilde{\mathbf{W}}_{HM}=\widetilde{\mathbf{W}}_{MH}$, $\widetilde{\mathbf{W}}_{MU}=\widetilde{\mathbf{W}}_{UM}$, and $\widetilde{\mathbf{W}}_{MU}=\widetilde{\mathbf{W}}_{UM}$), we only need to update three of out nine parts, namely $\widetilde{\mathbf{W}}_{HU}$, $\widetilde{\mathbf{W}}_{MU}$, and $\widetilde{\mathbf{W}}_{UU}$. The optimization problems of $\widetilde{\mathbf{W}}_{HU}$ and $\widetilde{\mathbf{W}}_{MU}$ can be formulated by
\begin{equation}
\label{eq10}
\begin{aligned}
       \mathop{\min}_{\widetilde{\mathbf{W}}_{HU(MU)}}\; \frac{\beta}{4}\norm{\widetilde{\mathbf{W}}\odot\mathbf{Z}}_{1,1} \; \mathrm{s.t.} \;  1/N_{k} \succeq \widetilde{\mathbf{W}}_{i,j} \succeq 0 ,\; \norm{\widetilde{\mathbf{W}}}_{1,1}=s,
\end{aligned}
\end{equation}
which can be solved by ADMM. More details can be found in \textbf{Algorithm~3}, where $\mathbf{Z}_{H(M)}$ and $\mathbf{Z}_{U}$ represent respectively the subspace features of $\mathbf{X}_{H(M)}$ and $\mathbf{X}_{U}$, $\prox$ stands for the proximal operator for $\norm{\widetilde{\mathbf{W}}}_{1,1}=s$ \cite{Heide:CVPR2015}. We technically add the constraint $\widetilde{\mathbf{W}}_{i,j} \preceq 1/N_{k}$ in order to share the same unit level with LDA-like graph.

For $\widetilde{\mathbf{W}}_{UU}$, the objective function can be written as
\begin{equation}
\label{eq11}
\begin{aligned}
   \mathop{\min}_{\widetilde{\mathbf{W}}_{UU}} \; \frac{\beta}{4}\norm{\widetilde{\mathbf{W}}\odot\mathbf{Z}}_{1,1} \;
   \mathrm{s.t.} \; \widetilde{\mathbf{W}}=\widetilde{\mathbf{W}}^{\T}, \; 1/N_k \succeq \widetilde{\mathbf{W}}_{i,j} \succeq 0 ,\; \norm{\widetilde{\mathbf{W}}}_{1,1}=s,
\end{aligned}
\end{equation}
which can be effectively solved using \textbf{Algorithm 4}.

Finally, we repeat these optimization procedures until a stopping criterion is satisfied.
\begin{figure}[!t]
	  \centering
		\subfigure[The University of Houston MS-HS Datasets]{
			\includegraphics[width=5.5cm,height=4cm]{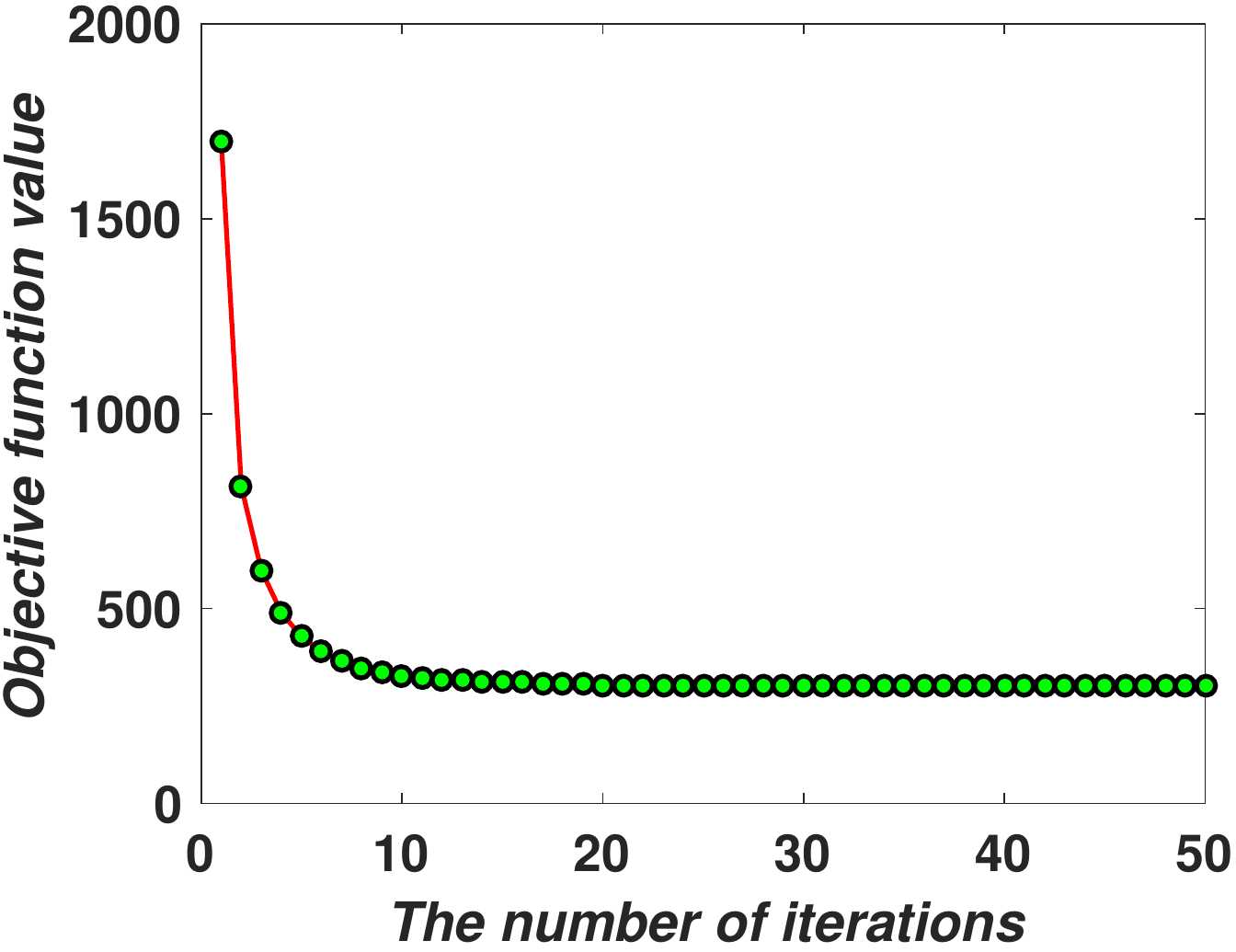}
            \label{fig:Convergence_HU}
		}
		\subfigure[The Chikusei MS-HS Datasets]{
			\includegraphics[width=5.5cm,height=4cm]{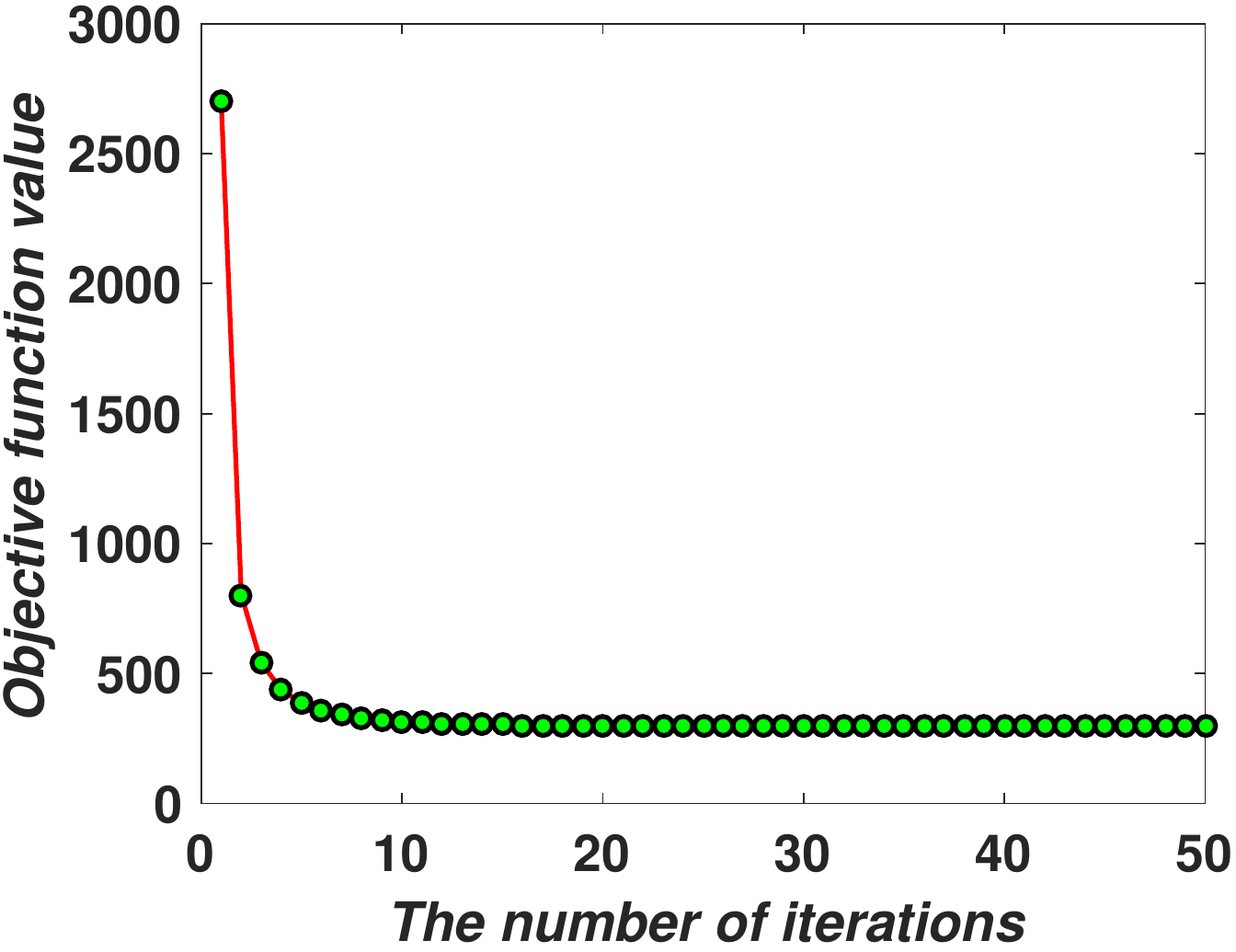}
            \label{fig:Convergence_CH}
		}
         \caption{Convergence analysis of LeMA are experimentally performed on the two MS-HS datasets.}
\label{fig:Convergence}
\end{figure}
\subsection{Convergence Analysis}
The alternative alternating strategy used in \textbf{Algorithm 1} is nothing but a block coordinate descent (BCD), which has been theoretically supported to converge to a stationary point as long as each subproblem in Eq. (\ref{eq4}) is exactly minimized~\cite{bertsekas1999nonlinear}. As observed, these subproblems with respect to the variables $\mathbf{P}$, $\mathbf{\Theta}$ and $\widetilde{\mathbf{W}}$ are strongly convex, and hence each independent task can ideally find an unique minimum when the Lagrangian parameter is updated within finitely iterative steps~\cite{boyd2011distributed}. Besides, ADMM used in each subproblem optimization is actually generalized to \emph{inexact} Augmented Lagrange Multiplier (ALM) \cite{chen2018equivalence}, whose convergence has been well studied when the number of block is less than three \cite{lin2010augmented} (e.g. \textbf{Algorithm 2}). Although there is still not a \emph{generally and strictly} theoretical proof in multi-blocks case, yet the convergence analysis for some common cases such as our \textbf{Algorithm 3} and \textbf{Algorithm 4} has been well conducted in \cite{Hong2017ICIP}\cite{liu2013robust}\cite{zhong2016blind}\cite{Zhou2017MultiADMM}. We also experimentally record the objective function values in each iteration to draw the convergence curves of LeMA on two used HS-MS datasets (see Fig. \ref{fig:Convergence}).
\section{Experiments}
In this section, we quantitatively and qualitatively evaluate the performance of the proposed method on two simulated HS-MS datasets (University of Houston and Chikusei) and a real multispectral-lidar and hyperspectral dataset provided by $2018$ IEEE GRSS data fusion contest (DFC2018), by the form of classification using two commonly used and high-performance classifiers, namely linear support vector machines (LSVM), and canonical correlation forest (CCF) ~\cite{CCF2015}. Three indices: overall accuracy (OA), average accuracy (AA), kappa coefficient ($\kappa$), are calculated to quantitatively assess the classification performance. Moreover, we compare the performance of the proposed LeMA and several other state-of-art algorithms, i.e. GLP ~\cite{zhu2003semi}, SMA, S-SMA~\cite{Wang:AAAI2009}, CoSpace and Semi-supervised CoSpace (S-CoSpace). The original MS data is used as a baseline. SMA constructs an LDA-like joint graph using label information. Besides label information, S-SMA method also uses unlabeled samples to generate the joint graph by computing the similarity based on Euclidean distance. The same strategy of graph construction is adopted for CoSpace and S-CoSpace.
\begin{table*}[!t]
\footnotesize
\centering
\caption{The number of training and testing samples for the two used MS-HS datasets.}
\smallskip
\begin{tabular}{c||c|c|c||c|c|c}
\hlinew{1.5pt}
\multirow{2}{*}{Class No.}&\multicolumn{3}{c||}{Houston MS-HS dataset}&\multicolumn{3}{c}{Chikusei MS-HS dataset}\\
\cline{2-7}&Class Name&Training&Testing&Class Name&Training&Testing\\
\hline \hline 1&Healthy Grass&537&699&Water&301&858\\
 2&Stressed Grass&61&1154&Bare Soil (School)&992&1867\\
 3&Synthetic Grass&340&357&Bare Soil (Farmland)&455&4397\\
 4&Tree&209&1035&Natural Plants&150&4272\\
 5&Soil&74&1168&Weeds in Farmland&928&1108\\
 6&Water&22&303&Forest&486&11904\\
 7&Residential&52&1203&Grass&989&5526\\
 8&Commercial&320&924&Rice Field (Grown)&813&8816\\
 9&Road&76&1149&Rice Field (First Stage)&667&1268\\
 10&Highway&279&948&Row Crops&377&5961\\
 11&Railway&33&1185&Plastic House&165&475\\
 12&Parking Lot1&329&904&Manmade (Non-dark)&170&568\\
 13&Parking Lot2&20&449&Manmade (Dark)&1291&6373\\
 14&Tennis Court&266&162&Manmade (Blue)&111&431\\
 15&Running Track&279&381&Manmade (Red)&35&187\\
 16&/&/&/&Manmade Grass&21&1019\\
 17&/&/&/&Asphalt&384&417\\
\hline \hline &Total&2897&12021&Total&8335&55447\\
\hlinew{1.5pt}
\end{tabular}
\label{Table:H_CH}
\end{table*}
\subsection{The Simulated MS-HS Datasets over the University of Houston}
\subsubsection{Data Description}
The HS data in the simulated \emph{Houston MS-HS datasets} was acquired by the ITRES-CASI-1500 sensor with the size of $349\times1905$ at a ground sampling distance (GSD) of 2.5m over the University of Houston campus and its neighboring urban areas. This data was provided for the $2013$ IEEE GRSS data fusion contest, with 144 bands covering the wavelength range from 364nm to 1046nm. Spectral simulation is performed to generate the MS image by degrading the HS image in the spectral domain using the MS spectral response functions (SRFs) of Sentinel-2 as filters (for more details refer to ~\cite{Naoto}). The MS data we used is generated with dimensions of $349\times1905\times10$.
\begin{figure}[!t]
	  \centering
		\subfigure{
			\includegraphics[width=0.9\textwidth]{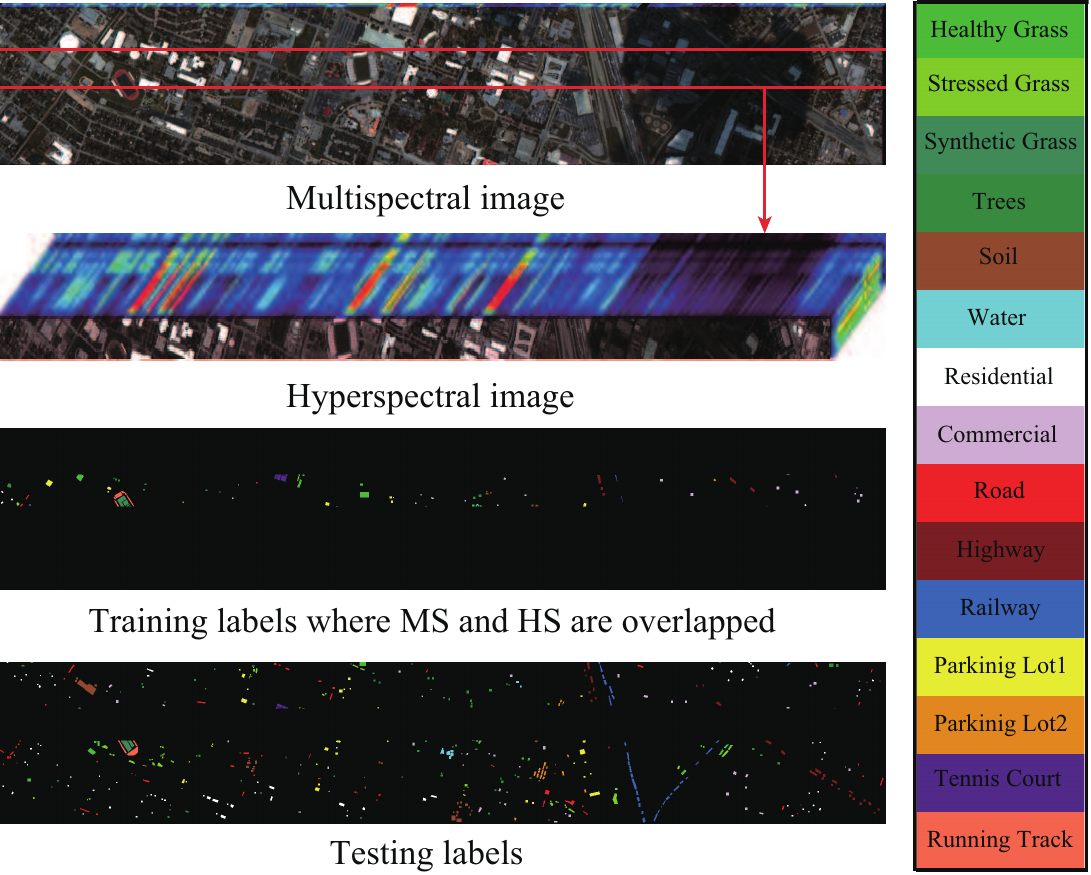}
		}
       \caption{The multispectral image and its corresponding hyperspectral image that partially covers the same area, as well as training and testing labels, for University of Houston dataset.}
\label{fig:Data_H}
\end{figure}
\subsubsection{Experimental Setup}
To meet our problem setting, a HS image partially overlapping MS image and a whole MS image are used in our experiments, and meanwhile the corresponding training and test samples can be re-assigned, as shown in Fig.~\ref{fig:Data_H}. In detail, since the total labels are available, we seek out a region where all kinds of classes are involved. The labels in the region are selected as the training set and the rest are seen as the test set, as shown in Fig.~\ref{fig:Data_H} and specifically quantified in Table \ref{Table:H_CH}.

The parameters of the different methods are determined by a 10-fold cross-validation on the training data. More specifically, we tune the parameters of the different algorithms to maximize their performances, e.g. dimension ($d$), penalty parameters ($\alpha, \beta$), etc. The dimension ($d$) is a common parameter for all compared algorithms, and it can be determined covering the range from $10$ to $50$ at an interval of $10$. For the number of nearest neighbors ($k$) and the standard deviation of Gaussian kernel function ($\sigma$) in artificially computing the adjacency matrix ($\mathbf{W}$) of GLP, SMA, and S-SMA, we select them in the range of $\{10, 20,...,50\}$ and $\{10^{-2}, 10^{-1}, \allowbreak 10^{0}, 10^{1}, 10^{2}\}$, respectively, Similarly to CoSpace, S-CoSpace and LeMA, we set the two regularization parameters ($\alpha, \beta$) ranging from $\{10^{-2}, 10^{-1}, \allowbreak 10^{0}, 10^{1}, 10^{2}\}$.
\subsubsection{Results and Analysis}
\begin{figure*}[!t]
	  \centering
		\subfigure{
			\includegraphics[width=0.9\textwidth]{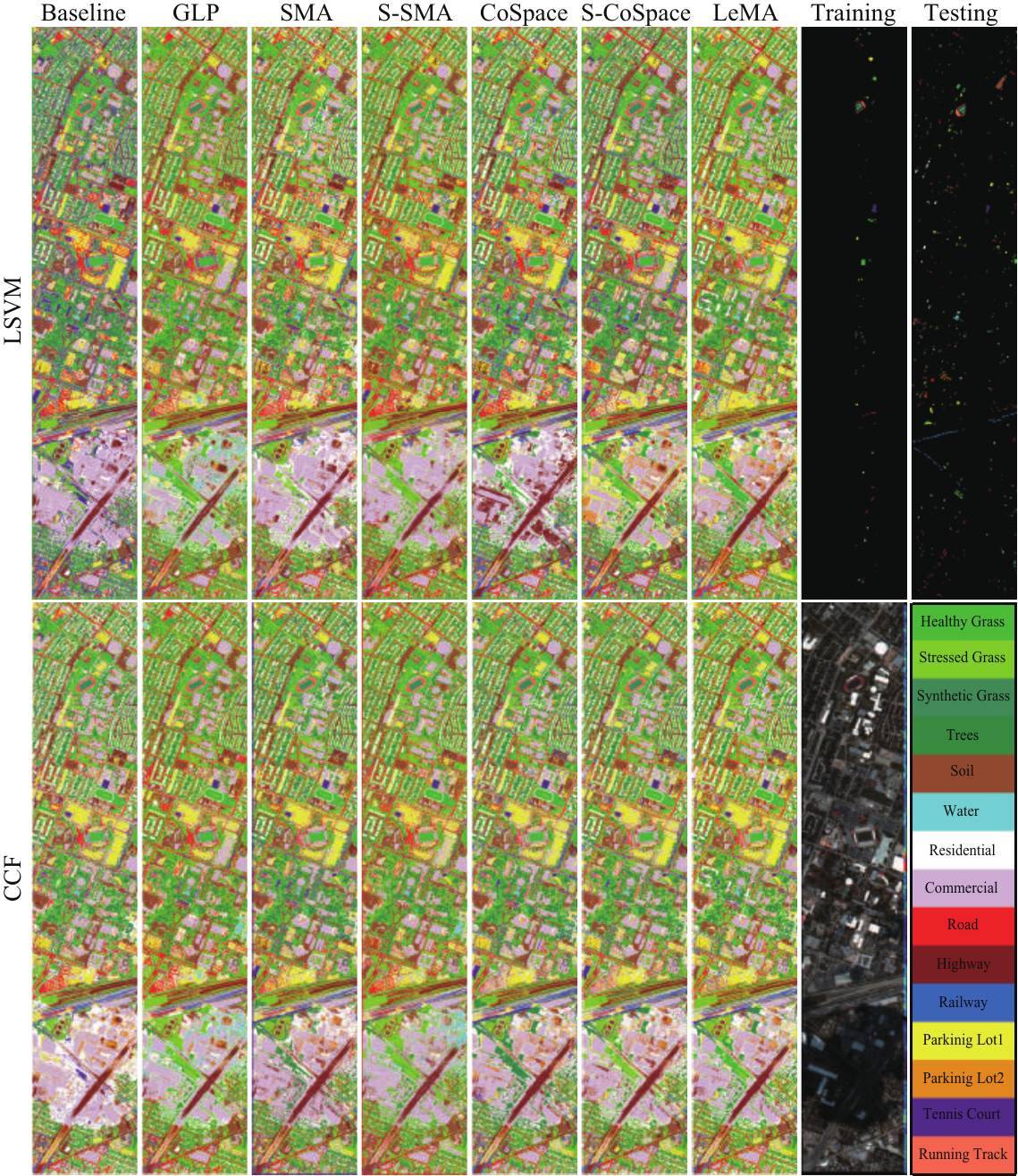}
		}
        \caption{Classification maps of the different algorithms obtained using two kinds of classifiers on the University of Houston dataset.}
\label{fig:ClassificationMap_H}
\end{figure*}

Fig.\ref{fig:ClassificationMap_H} shows the classification maps of compared algorithms using LSVM and CCF classifiers, while Table~\ref{tab:Houston} lists the specific quantitative assessment results with optimal parameters obtained by 10-fold cross-validation.
\begin{table*}[!t]
\centering
\caption{Quantitative performance comparison with the different algorithms on the University of Houston data. The best one is shown in bold.}
\resizebox{\textwidth}{!}{
\begin{tabular}{c||c|c||c|c||c|c||c|c||c|c||c|c||c|c}
\hlinew{1.5pt}Methods&\multicolumn{2}{c||}{Baseline (\%) }&\multicolumn{2}{c||}{GLP (\%) }&\multicolumn{2}{c||}{SMA (\%) }&\multicolumn{2}{c||}{S-SMA (\%) }&\multicolumn{2}{c||}{CoSpace (\%) }&\multicolumn{2}{c||}{S-CoSpace (\%) }&\multicolumn{2}{c}{LeMA (\%) }\\
\hline \hline \multirow{2}{*}{Parameter}&\multicolumn{2}{c||}{$d$}&\multicolumn{2}{c||}{$(k,\sigma,d)$}&\multicolumn{2}{c||}{$d$}&\multicolumn{2}{c||}{$(k,\sigma,d)$}&\multicolumn{2}{c||}{$(\alpha,\beta,d)$}&\multicolumn{2}{c||}{$(\alpha,\beta,d)$}&\multicolumn{2}{c}{$(\alpha,\beta,d)$}\\
\cline{2-15} &\multicolumn{2}{c||}{10}&\multicolumn{2}{c||}{$(10,1,10)$}&\multicolumn{2}{c||}{$30$}&\multicolumn{2}{c||}{$(10,0.1,30)$}&\multicolumn{2}{c||}{$(0.01,0.01,30)$}&\multicolumn{2}{c||}{$(0.1,0.01,30)$}&\multicolumn{2}{c}{$(0.01,0.01,30)$}\\
\hline \hline Classifier& LSVM & CCF & LSVM & CCF & LSVM & CCF& LSVM & CCF & LSVM& CCF & LSVM & CCF& LSVM & CCF \\
\hline \hline OA&62.12&68.21&64.71&70.01&68.01&69.59&69.29&70.10&69.38&72.17&70.41&73.75&73.42&\textbf{76.35}\\
              AA&65.97&70.47&68.18&72.18&70.50&71.02&72.00&72.88&71.69&73.56&73.12&75.61&74.76&\textbf{77.18}\\
              $\kappa$&0.5889&0.6543&0.6164&0.6728&0.6520&0.6695&0.6659&0.6754&0.6672&0.6975&0.6784&0.7146&0.7110&\textbf{0.7428}\\
\hline \hline Class1&76.39&67.95&77.83&77.97&75.25&68.53&74.25&73.53&75.54&69.96&\textbf{91.85}&87.98&89.56&85.84\\
              Class2&80.59&78.08&93.85&\textbf{98.01}&97.57&77.9&97.57&93.67&73.74&77.99&90.12&91.59&93.67&93.85\\
              Class3&\textbf{100.00}&\textbf{100.00}&\textbf{100.00}&\textbf{100.00}&\textbf{100.00}&\textbf{100.00}&\textbf{100.00}&\textbf{100.00}&\textbf{100.00}&\textbf{100.00}&\textbf{100.00}&\textbf{100.00}&\textbf{100.00}&\textbf{100.00}\\
              Class4&85.51&92.27&89.66&96.62&94.78&98.74&95.85&98.55&98.74&98.26&92.75&97.29&97.49&\textbf{99.61}\\
              Class5&99.06&99.4&99.49&\textbf{99.66}&98.97&99.14&99.32&99.4&99.4&99.4&99.4&\textbf{99.66}&99.49&99.57\\
              Class6&86.14&86.14&96.37&99.01&86.47&70.96&\textbf{99.67}&99.67&85.48&85.15&\textbf{99.67}&96.70&86.47&86.47\\
              Class7&50.62&63.76&48.63&64.01&72.32&77.14&72.15&69.66&73.98&80.05&75.06&80.96&83.21&\textbf{88.03}\\
              Class8&56.49&56.06&56.60&59.85&62.01&62.23&\textbf{64.61}&63.85&63.53&62.01&55.84&60.39&62.77&62.01\\
              Class9&56.22&70.58&69.63&69.02&49.96&61.27&50.57&45.00&59.79&64.93&65.8&\textbf{71.54}&64.49&61.88\\
              Class10&45.36&45.25&45.46&49.89&58.12&52.32&58.33&63.61&\textbf{64.14}&57.70&58.97&51.79&60.97&53.59\\
              Class11&27.43&43.88&22.45&38.65&28.86&36.46&36.46&34.77&36.54&47.26&35.78&38.65&41.27&\textbf{49.96}\\
              Class12&31.64&56.08&31.75&37.83&35.84&62.50&34.18&55.2&46.79&62.72&34.29&58.52&45.02&\textbf{76.88}\\
              Class13&0.00&0.67&0.00&1.11&0.00&0.00&0.00&0.45&0.00&0.45&0.00&0.89&0.00&\textbf{1.78}\\
              Class14&97.53&98.77&94.44&92.59&\textbf{100.00}&\textbf{100.00}&99.38&98.15&\textbf{100.00}&99.38&99.38&\textbf{100.00}&99.38&\textbf{100.00}\\
              Class15&96.59&98.16&96.59&\textbf{98.43}&97.38&98.16&97.64&97.64&97.64&98.16&97.90&98.16&97.64&98.16\\
\hlinew{1.5pt}
\end{tabular}}
\label{tab:Houston}
\end{table*}

Overall, the methods based on manifold alignment outperform baseline and GLP using the different classifiers. This means that the limited amount of HS data can guide the corresponding MS data towards better discriminative feature representations. More specifically when compared with S-SMA, SMA yields a relatively poor performance since it only considers the correspondences of MS-HS labeled data. This indicates that reasonably embedding unlabeled samples into the manifold alignment framework can effectively help us capture the real data distribution, and thereby obtain more accurate decision boundaries. Unfortunately, these approaches only attempt to align different data in a common subspace, but they hardly take the connections between the common subspace and label information into account\footnote{The connectivity in manifold alignment is not strictly equivalent to the similarity of the two samples.}, which leads to a lack of discriminative ability. With regards to this, our proposed joint learning framework ``CoSpace'' and its semi-supervised version ``S-CoSpace'' achieve the desired results on the the given MS-HS datasets.

By fully considering the connectivity of the common subspace, label information, and unlabeled information encoded by the learned graph structure, the performance of LeMA is much more superior to that of any other methods as can be observed in Table~\ref{tab:Houston}. This demonstrates that LeMA is likely to learn a more discriminative feature representation and to find a better decision boundary.

As observed from Fig. \ref{fig:Data_H} and Table \ref{tab:Houston}, the training samples are relatively a few and meanwhile the distribution between different classes is extremely unbalanced. While training the classifier, more attentions are paid on those classes with large-size samples, and some small-scale classes possibly play less and even nothing. For this reason, we propose to consider those large-scale unlabeled data, achieving a semi-supervised learning. Using this strategy, the semi-supervised methods, i.e. GLP, S-SMA, S-CoSpace, obviously perform better than baseline and their supervised ones (SMA and CoSpace). Moreover, we can see from Table \ref{tab:Houston} that there is a significant improvement of classification performance in some classes (e.g.\emph{ Stressed Grass}, \emph{Water}) after accounting for unlabeled samples, particularly between SMA and S-SMA as well as CoSpace and S-CoSpace. However, these aforementioned semi-supervised methods carry out the label propagation on a given graph manually computed by gaussian kernel function, limiting the adaptiveness and discriminability of the algorithms. LeMA can adaptively learn a data-driven graph structure where the labels tend to spread more smoothly, which can result in a more effective material identification for those challenging classes (few training samples), such as \emph{Trees}, \emph{Residential}, \emph{Railway}, \emph{Parking Lot1}. In addition, we can also observe an easily overlooked phenomenon that the LeMA's ability in identifying certain classes still remains limited, such as \textit{Parking Lot2}(only $1.78\%$) and \textit{Railway} ($49.96\%$). \textit{Parking Lot2} is basically classified to \textit{Commercial} and \textit{Parking Lot1}, while \textit{Railway} is largely identified as \textit{Road} and \textit{Commercial}. This might be explained by the limited number of training samples as well as fairly similar spectral properties between several classes.

\subsection{The Simulated MS-HS Datasets over Chikusei}
\subsubsection{Data Description}
Similarly to Houston data, the MS data with dimensions of $2517\times2335\times10$ at a GSD of 2.5 m was simulated by the HS data acquired by the Headwall$'$s Hyperspec-VNIR-C sensor over Chikusei area, Ibaraki, Japan. It consists of 128 bands in the spectral range from 363nm to 1018nm with the 10nm spectral resolution. The dataset has been made available to the scientific research~\cite{yokoya2016airborne}.
\begin{figure}[!t]
	  \centering
		\subfigure{
			\includegraphics[width=0.9\textwidth]{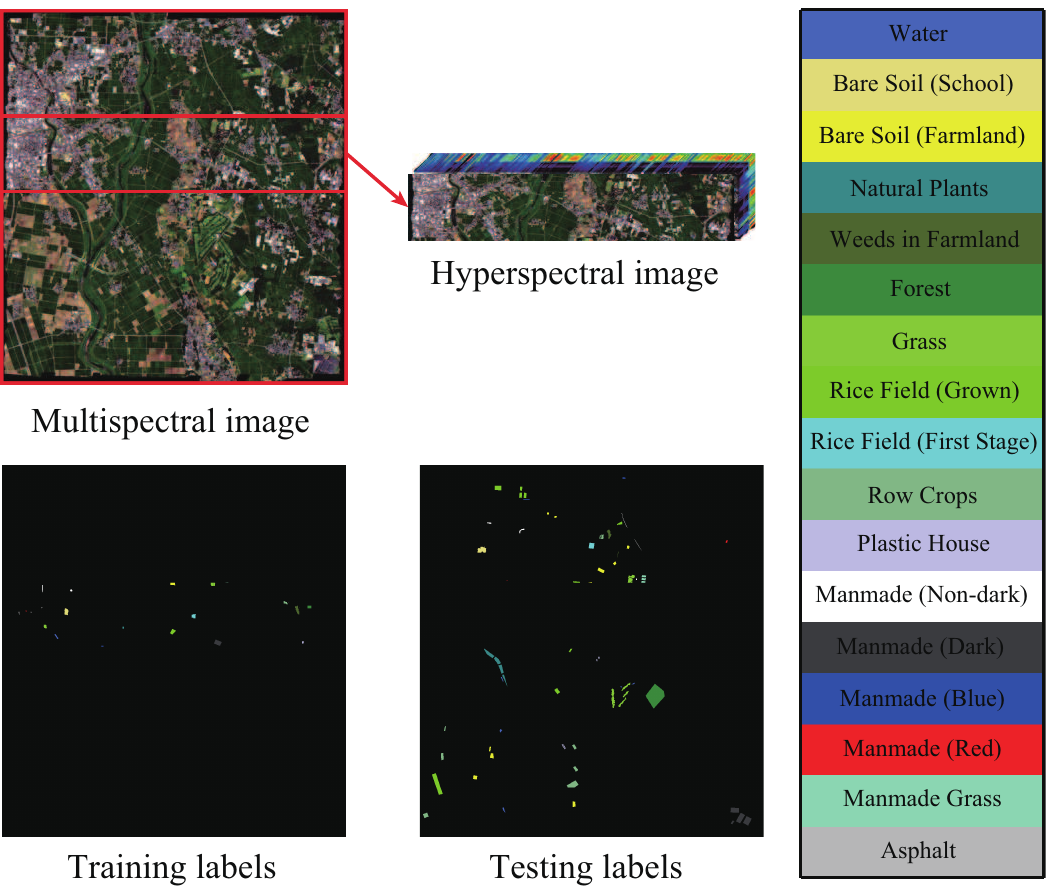}
		}
       \caption{The multispectral image and its corresponding hyperspectral image that partially covers the same area, as well as training and testing labels, for Chikusei Dataset.}
\label{fig:Data_CH}
\end{figure}

\begin{figure*}[!t]
	  \centering
		\subfigure{
			\includegraphics[width=0.9\textwidth]{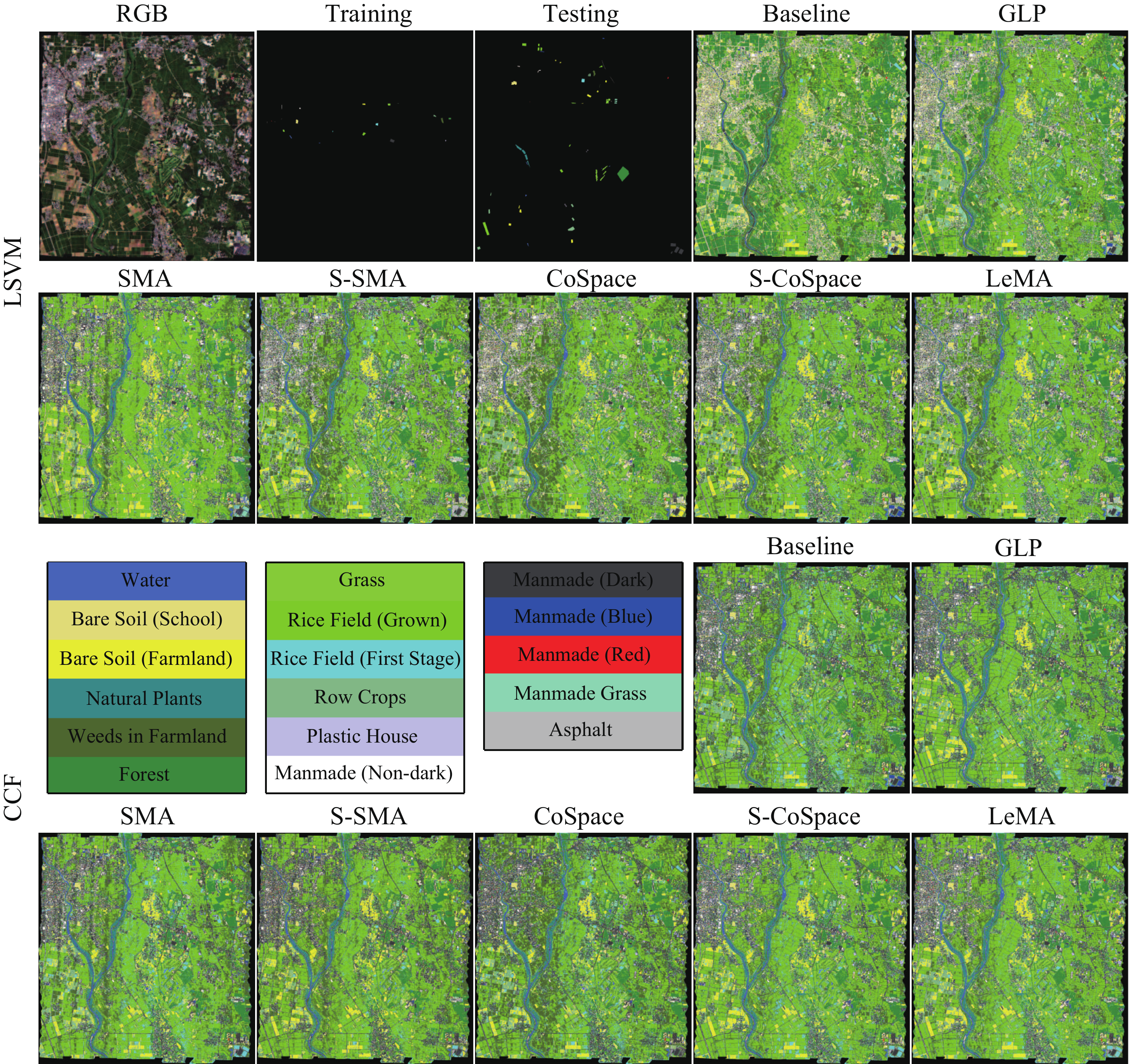}
		}
        \caption{Classification maps of the different algorithms obtained using two kinds of classifiers on the Chikusei dataset.}
\label{fig:ClassificationMap_CH}
\end{figure*}
\subsubsection{Experimental Setup}
Fig.~\ref{fig:Data_CH} shows the corresponding MS and partial HS images as well as selected training labels and test labels. Again, the overlapped region between MS and HS, which should include all the classes listed in Table \ref{Table:H_CH}, is chosen based on the given ground truth~\cite{yokoya2016airborne}. Additionally, the parameters configuration for all algorithms can be adaptively completed by a 10-fold cross-validation on the training set, which is more generalized to different datasets. Regarding how to run the cross-validation for parameters setting, please refer to section 3.1.2 for more details.
\subsubsection{Results and Analysis}
We assess the classification performance of the different algorithms for the Chikusei MS-HS data both quantitatively and visually, as shown in Fig.\ref{fig:ClassificationMap_CH} and Table~\ref{tab:Chikusei}.
\begin{table*}[!t]
\centering
\caption{Quantitative performance comparison with the different algorithms on the Chikusei data. The best one is shown in bold.}
\resizebox{\textwidth}{!}{
\begin{tabular}{c||c|c||c|c||c|c||c|c||c|c||c|c||c|c}
\hlinew{1.5pt}Methods&\multicolumn{2}{c||}{Baseline (\%) }&\multicolumn{2}{c||}{GLP (\%) }&\multicolumn{2}{c||}{SMA (\%) }&\multicolumn{2}{c||}{S-SMA (\%) }&\multicolumn{2}{c||}{CoSpace (\%) }&\multicolumn{2}{c||}{S-CoSpace (\%) }&\multicolumn{2}{c}{LeMA (\%) }\\
\hline \hline \multirow{2}{*}{Parameter}&\multicolumn{2}{c||}{$d$}&\multicolumn{2}{c||}{$(k,\sigma,d)$}&\multicolumn{2}{c||}{$d$}&\multicolumn{2}{c||}{$(k,\sigma,d)$}&\multicolumn{2}{c||}{$(\alpha,\beta,d)$}&\multicolumn{2}{c||}{$(\alpha,\beta,d)$}&\multicolumn{2}{c}{$(\alpha,\beta,d)$}\\
\cline{2-15} &\multicolumn{2}{c||}{10}&\multicolumn{2}{c||}{$(10,1,10)$}&\multicolumn{2}{c||}{$20$}&\multicolumn{2}{c||}{$(10,0.1,20)$}&\multicolumn{2}{c||}{$(0.1,0.01,30)$}&\multicolumn{2}{c||}{$(0.1,0.01,30)$}&\multicolumn{2}{c}{$(0.1,0.01,30)$}\\
\hline \hline Classifier& LSVM & CCF & LSVM & CCF & LSVM & CCF& LSVM & CCF & LSVM& CCF & LSVM & CCF& LSVM & CCF \\
\hline \hline OA&60.20&71.11&62.30&72.26&67.90&71.53&69.68&73.27&71.12&75.69&72.60&77.11&75.11&\textbf{81.71}\\
              AA&69.42&70.40&69.80&70.71&70.79&66.47&72.27&70.01&73.96&71.46&71.64&71.33&75.29&\textbf{75.73}\\
              $\kappa$&0.5523&0.6761&0.5784&0.6894&0.6391&0.6802&0.6602&0.6818&0.6746&0.7260&0.6911&0.7420&0.7194&\textbf{0.7933}\\
\hline \hline Class1&78.21&80.54&78.09&80.42&98.72&82.52&\textbf{99.53}&97.90&92.54&79.25&98.83&98.37&98.25&98.83\\
              Class2&94.43&82.70&94.11&93.84&93.20&92.50&93.20&93.09&93.47&\textbf{94.91}&87.04&93.63&93.20&93.79\\
              Class3&23.54&50.06&37.75&76.87&62.57&55.31&68.41&76.55&80.40&77.71&80.65&77.23&89.29&\textbf{89.90}\\
              Class4&92.13&92.56&92.23&95.72&90.57&91.53&92.51&88.76&90.59&96.23&94.64&92.49&95.11&\textbf{96.96}\\
              Class5&\textbf{97.65}&94.68&96.84&88.45&28.43&16.06&24.01&32.85&83.94&66.52&51.81&43.32&60.74&67.78\\
              Class6&62.01&81.48&57.47&69.67&62.52&78.91&68.27&79.67&63.61&79.02&72.34&\textbf{88.48}&76.34&87.27\\
              Class7&99.67&99.93&99.66&\textbf{100.00}&96.87&97.79&95.40&99.37&97.74&99.75&98.41&99.87&97.63&99.80\\
              Class8&57.11&93.40&69.06&98.93&95.59&93.49&96.88&96.53&95.05&92.72&\textbf{99.48}&98.45&99.27&99.18\\
              Class9&\textbf{100.00}&\textbf{100.00}&\textbf{100.00}&99.92&99.53&99.13&99.45&99.21&98.66&99.76&99.21&98.34&99.76&\textbf{100.00}\\
              Class10&24.81&19.56&\textbf{26.64}&19.06&21.39&15.48&20.94&13.09&22.35&18.00&22.75&14.83&26.47&26.46\\
              Class11&0.00&2.11&0.00&0.00&0.00&0.00&0.00&0.00&0.00&0.00&0.21&5.47&0.63&\textbf{5.68}\\
              Class12&\textbf{90.32}&88.91&\textbf{90.32}&89.61&90.14&85.92&90.14&89.44&90.32&80.46&89.96&89.44&88.38&90.14\\
              Class13&33.11&33.09&33.11&36.50&32.61&56.25&31.32&30.88&33.11&67.90&33.11&54.93&33.11&\textbf{68.73}\\
              Class14&\textbf{94.20}&85.38&79.12&59.40&72.85&59.40&\textbf{94.20}&86.31&59.40&52.44&14.39&49.19&45.01&53.60\\
              Class15&\textbf{100.00}&\textbf{100.00}&\textbf{100.00}&\textbf{100.00}&93.58&\textbf{100.00}&\textbf{100.00}&\textbf{100.00}&93.58&97.86&\textbf{100.00}&\textbf{100.00}&\textbf{100.00}&\textbf{100.00}\\
              Class16&74.88&88.62&74.19&93.52&99.71&99.51&99.80&98.82&97.84&\textbf{100.00}&97.35&97.25&98.04&95.78\\
              Class17&58.03&3.84&58.03&0.24&65.23&7.91&62.11&7.67&64.75&0.00&77.70&11.27&\textbf{78.66}&13.43\\
\hlinew{1.5pt}
\end{tabular}}
\label{tab:Chikusei}
\end{table*}

Similarly to the University of Houston MS-HS data, there is a basically consistent trend for the different algorithms in the Chikusei MS-HS data. On the whole, the original MS data (baseline) fails to identify some specific materials such as \emph{Plastic House}, \emph{Manmade (Dark)}, \emph{Rice Field (Grown)}, \emph{Bare Soil (Farmland)}, and \emph{Forest}, due to its poor spectral information and a limited number of training samples. GLP utilizes the unlabeled samples to augment the training samples in a semi-supervised way, yet it is still limited by the low-discriminative spectral signatures. By aligning the MS and HS data, these alignment-based approaches (e.g. SMA, S-SMA, CoSpace, S-CoSpace, and LeMA) are able to find a common subspace in which the learnt features are expected to absorb the different properties from two modalities, resulting in a better performance. Compared to the supervised methods (SMA and CoSpace), their corresponding semi-supervised versions (S-SMA and S-CoSpace) obtain higher classification accuracies on both classifiers, which is detailed in Table~\ref{tab:Chikusei}. As expected, the performance of the LeMA is significantly superior to that of others, thanks to the great contributions of a common subspace learning from MS-HS data, a data-driven graph learning and the semi-supervised learning strategy. Despite so, the LeMA still fails to recognize some challenging classes, such as \textit{Weeds in Farmland}, \textit{Row Crops}, \textit{Plastic House}, and \textit{Asphalt}. The reasons could be two-fold. On one hand, the performance of LeMA is limited, to some extent, by the unbalanced data sets. On the other hand, LeMA' transferring ability would sharply degrade when a great spectral variability between training and test samples exists.
\begin{figure*}[!t]
	  \centering
		\subfigure{
			\includegraphics[width=0.9\textwidth]{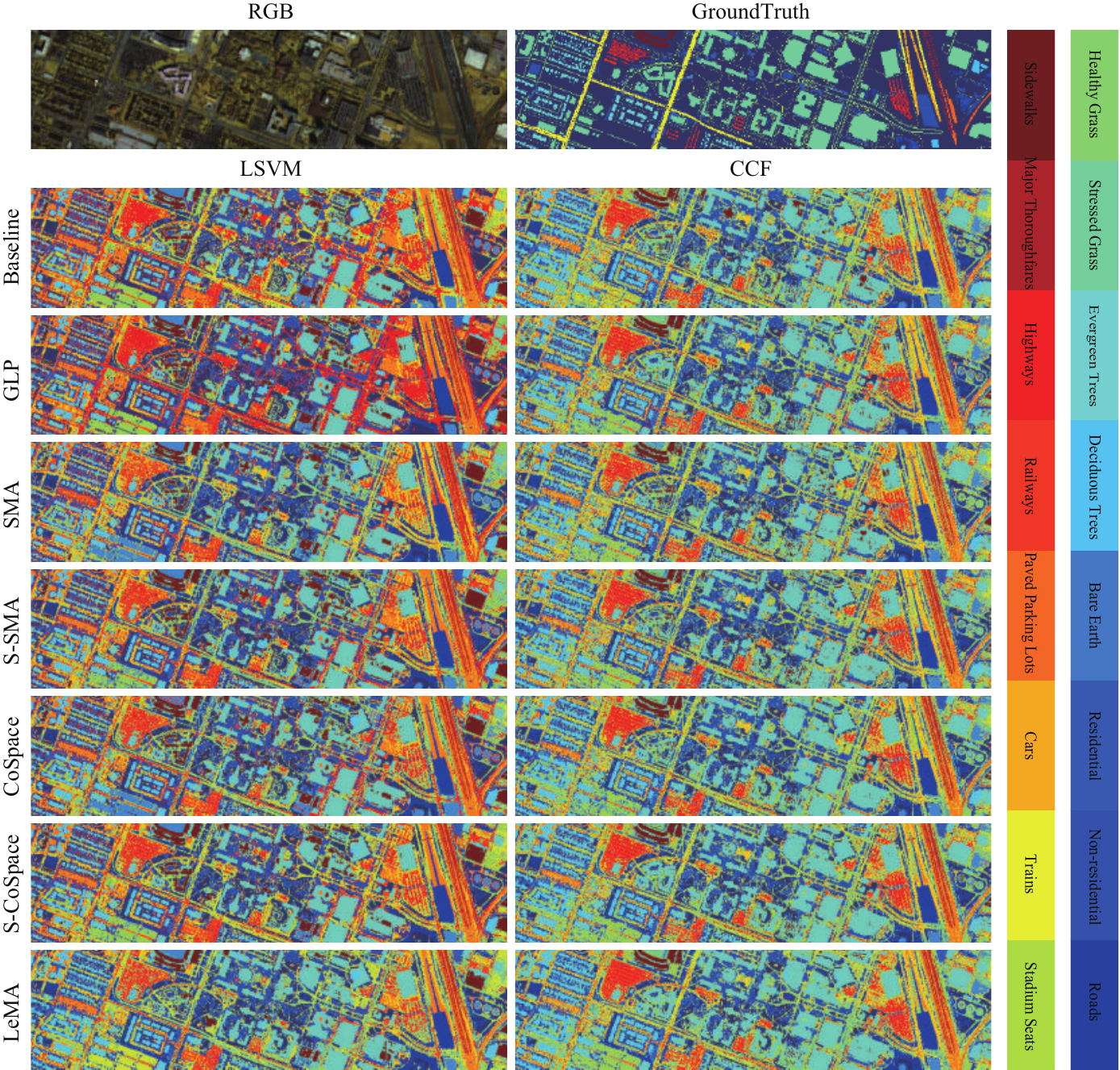}
		}
        \caption{Classification maps of the different algorithms obtained using two kinds of classifiers on the real dataset of DFC2018 (Multispectral-Lidar and Hyperspectral data).}
\label{fig:ClassificationMap_DFC2018}
\end{figure*}

\subsection{The Real Multispectral-Lidar and Hyperspectral Datasets in DFC2018}
Although we follow strict simulation procedures, yet the two MS-HS datasets used above (Houston and Chikusei) essentially originate from a similar data source (homogeneous), which means there is a strong correlation in their spectral features. This makes the information of the different modalities transferred more effectively, but could limit the generalization ability in practice. To this end, we apply a real bi-modal dataset -- multispectral-lidar and hyperspectral (heterogeneous) provided by the latest IEEE GRSS data fusion contest 2018 (DFC2018).
\subsubsection{Data Description}
Multi-source optical remote sensing data, such as multispectral-lidar data, hyperspectral data, and very high-resolution RGB data, is provided in the contest. More specifically, the multispectral-lidar imagery consists of $1202\times4768$ pixels with 7 bands ( 3 intensity bands and 4 DSMs-related bands \cite{le20182018}) collected from 1550nm, 1064nm, and 532nm at a 0.5m GSD, while the hyperspectral data comprises 48 bands covering a spectral range from 380nm to 1050nm at 1m GSD, and its size is $601\times2384$. In our case, our LeMA model is trained on partial multispectral-lidar and hyperspectral correspondences and tested only using multispectral-lidar data, in order to meet the requirement of our cross-modality learning task. The first row of Fig.\ref{fig:ClassificationMap_DFC2018} shows the RGB image of this scene and the labeled ground truth image.
\begin{table*}[!t]
\centering
\caption{Quantitative performance comparison with the different algorithms on the DFC2018 data. The best one is shown in bold.}
\resizebox{\textwidth}{!}{
\begin{tabular}{c||c|c||c|c||c|c||c|c||c|c||c|c||c|c}
\hlinew{1.5pt}Methods&\multicolumn{2}{c||}{Baseline (\%) }&\multicolumn{2}{c||}{GLP (\%) }&\multicolumn{2}{c||}{SMA (\%) }&\multicolumn{2}{c||}{S-SMA (\%) }&\multicolumn{2}{c||}{CoSpace (\%) }&\multicolumn{2}{c||}{S-CoSpace (\%) }&\multicolumn{2}{c}{LeMA (\%) }\\
\hline \hline \multirow{2}{*}{Parameter}&\multicolumn{2}{c||}{$d$}&\multicolumn{2}{c||}{$(k,\sigma,d)$}&\multicolumn{2}{c||}{$d$}&\multicolumn{2}{c||}{$(k,\sigma,d)$}&\multicolumn{2}{c||}{$(\alpha,\beta,d)$}&\multicolumn{2}{c||}{$(\alpha,\beta,d)$}&\multicolumn{2}{c}{$(\alpha,\beta,d)$}\\
\cline{2-15} &\multicolumn{2}{c||}{7}&\multicolumn{2}{c||}{$(10,1,7)$}&\multicolumn{2}{c||}{$30$}&\multicolumn{2}{c||}{$(10,1,30)$}&\multicolumn{2}{c||}{$(0.1,0.1,30)$}&\multicolumn{2}{c||}{$(0.1,0.01,30)$}&\multicolumn{2}{c}{$(0.1,0.01,30)$}\\
\hline \hline Classifier& LSVM & CCF & LSVM & CCF & LSVM & CCF& LSVM & CCF & LSVM& CCF & LSVM & CCF& LSVM & CCF \\
\hline \hline OA&51.35&72.84&52.28&73.15&52.73&70.37&54.69&72.13&55.56&74.04&58.65&76.59	&61.69&\textbf{79.98}
\\
              AA&59.46&78.64&60.57&81.64&58.06&77.78&65.34&78.72&66.16&80.46&67.72&83.67&65.54&\textbf{88.82}
\\
              $\kappa$&0.4194&0.6534&0.4289&0.6587&0.4366&0.6256&0.4598&0.6441&0.4670&0.6682&0.4987&0.6990&0.5284&\textbf{0.7414}
\\
\hline \hline Class1&91.70&84.62&96.15&93.12&84.01&85.43&94.13&90.89&95.14&89.07&94.74&95.14&92.31&\textbf{100.00}
\\
              Class2&33.90&80.17&35.62&80.74&73.00&82.40&69.57&80.17&61.32&80.37&69.73&81.52&78.09&\textbf{87.90}
\\
              Class3&94.92&96.16&96.02&96.57&95.06&95.06&96.30&96.30&93.83&97.26&94.79&96.30&96.57&\textbf{99.45}
\\
              Class4&83.00&92.50&85.50&97.50&85.50&90.00&84.50&94.00&83.00&91.00&85.50&98.00&79.00&\textbf{100.00}
\\
              Class5&43.71&90.42&30.54&87.43&53.29&87.43&52.10&85.03&61.08&92.22&45.51&92.22&30.54&\textbf{100.00}
\\
              Class6&80.44&90.60&81.32&91.82&78.79&87.77&82.80&87.98&83.94&90.35&85.24&91.27&89.71&\textbf{96.50}
\\
              Class7&59.26&82.01&61.11&81.52&57.62&78.21&58.66&82.45&59.89&82.37&63.95&85.14&69.56&\textbf{87.47}
\\
              Class8&14.07&31.98&10.75&36.00&21.71&	28.00&20.83&35.16&26.64&38.71&11.77&39.51&31.43&\textbf{49.96}
\\
              Class9&48.54&54.14&50.77&58.40&44.87&56.96&52.60&53.49&47.94&63.30&53.69&\textbf{68.55}&40.47&62.26
\\
              Class10&10.16&42.07&8.00&31.70&6.77&37.82&5.55&29.21&11.02&36.67&24.21&\textbf{38.40}&12.93&38.04
\\
              Class11&23.54&72.03&25.96&79.07&79.07&74.45&45.88&75.45&34.21&76.26&54.12&81.49&62.58&\textbf{100.00}
\\
              Class12&93.85&85.85&92.92&94.46&92.00&87.08&85.85&90.15&85.54&86.15&74.15&95.38&66.46&\textbf{100.00}
\\
              Class13&60.50&74.96&57.31&87.56&59.33&73.45&60.17&77.98&63.03&79.33&64.71&87.06&70.59&\textbf{99.83}
\\
              Class14&39.93&87.15&55.21&90.63&17.71&86.11&47.22&85.76&66.32&89.58&75.69&90.63&55.21&\textbf{99.65}
\\
              Class15&95.39&96.77&97.70&\textbf{100.00}&93.55&98.16&99.54&97.70&99.54&98.62&99.54&\textbf{100.00}&95.85&\textbf{100.00}
\\
              Class16&78.39&96.77&84.19&99.68&77.74&96.13&89.68&97.74&86.13&96.13&86.13&98.06&77.42&\textbf{100.00}
\\
\hlinew{1.5pt}
\end{tabular}}
\label{tab:DFC2018}
\end{table*}
\subsubsection{Experimental Setup}
Our aim is, once again, to investigate whether the limited amount of hyperspectral data can improve the performance of another modality, e.g., multispectral data (homogeneous) or multispectral-lidar data (heterogeneous). Therefore, we randomly assign $10\%$ of total labeled samples as training set and the rest of it as test set in the experiment. Moreover, 16 main classes are selected out of 20 (see Fig.\ref{fig:ClassificationMap_DFC2018}), by removing several small classes with too few samples, e.g. \emph{Artificial Turf}, \emph{Water}, \emph{Crosswalks}, and \emph{Unpaved Parking Lots}. Likewise, we automatically configure the parameters of the proposed LeMA and the compared algorithms by a 10-fold cross-validation on the training set, which is detailed in section 3.1.2.
\subsubsection{Results and Analysis}
We show the averaged results of the different algorithms out of 10 runs to obtain a relatively stable and meaningful performance comparison, because the training and test sets are randomly generated from total samples in each round, as listed in Table \ref{tab:DFC2018}. Correspondingly, Fig. \ref{fig:ClassificationMap_DFC2018} visually highlights the differences of classification maps for the different methods.

Generally speaking, hyperspectral information embedding can effectively improve the classification performance of the multispectral-lidar data, which implies that the models based common subspace learning (e.g., SMA, S-SMA, CoSpace, S-CoSpace, and LeMA) can transfer the knowledge from one modality to another modality to some extent. We also observe from Table \ref{tab:DFC2018} that the semi-supervised methods which consider the unlabeled samples (e.g., GLP, S-SMA, S-CoSpace, and LeMA) always perform better than those purely supervised ones. Not unexpectedly, LeMA integrating rich spectral information and unlabeled samples achieves a superior performance, which demonstrates that the learning-based graph structure is more applicable to capturing the data distribution and further find a potential optimal decision boundary.

One thing to be noted, however, is that compared to the performance of the different algorithms in the simulated MS-HS datasets from similar sources (homogeneous), the knowledge transferring ability of these algorithms in handling the real multispectral-lidar and hyperspectral datasets from different sources (heterogeneous) remains limited, since all listed methods including our LeMA are modeled in a linearized way. Unfortunately, a single linear transformation fails to fit the gap between heterogeneous modalities well, despite a limited performance improvement.

\section{Conclusions}
In real-world problems, a large amount of low-quality data (e.g. MS data) can often be easily collected. On the contrary, high-quality data (e.g. HS data) are usually expensive and difficult to obtain. This motivates us to investigate whether a limited amount of high-quality data can contribute to relevant tasks with a large amount of low-quality data. For this purpose, we propose a novel semi-supervised learning framework called LeMA, which effectively connects the common subspace and label information, and automatically embeds the unlabeled information into the proposed framework by adaptively learning a Laplacian matrix from the data. Extensive experiments are conducted using the LeMA on two homologous MS-HS simulated datasets and a heterogenous multispectral-lidar and hyperspectral real dataset in comparison with the other state-of-arts algorithms, demonstrating the superiority and effectiveness of the LeMA in the knowledge transferring ability. We have to admit, however, that despite a significant performance improvement in LeMA, yet its representative ability is still limited by linearly modeling way, especially facing highly-nonlinear heterogenous data. Towards this issue, we will continue to improve our model to a nonlinear version and simultaneously consider the spatial information (e.g., morphological profiles) to further strengthen the feature representation ability.
\section{Acknowledgements}
The authors would like to thank the Hyperspectral Image Analysis group and the NSF Funded Center for Airborne Laser Mapping (NCALM) at the University of Houston for providing the CASI University of Houston dataset. The authors would like to express their appreciation to Prof. D. Cai and Dr. C. Wang for providing MATLAB codes for LPP and manifold alignment algorithms.

This work was supported by funding from the European Research Council (ERC) under the European Union's Horizon 2020 research and innovation program (grant agreement No [ERC-2016-StG-714087]) and from Helmholtz Association under the framework of the Young Investigators Group ''SiPEO'' (VH-NG-1018, www.sipeo.bgu. tum.de). The work of N. Yokoya was supported by Japan Society for the Promotion of Science (JSPS) KAKENHI 15K20955 and Alexander von Humboldt Fellowship for postdoctoral researchers.

\bibliographystyle{elsarticle-num}
\bibliography{LeMA}





\end{document}